\documentclass[lettersize,journal]{IEEEtran}
\usepackage{amssymb}
\usepackage{amsmath,amsfonts}
\usepackage{algorithmic}
\usepackage{algorithm}
\usepackage{lineno}
\usepackage{array}
\usepackage{orcidlink}
\usepackage[caption=false,font=normalsize,labelfont=sf,textfont=sf]{subfig}
\usepackage{textcomp}
\usepackage{stfloats}
\usepackage{url}
\usepackage{verbatim}
\usepackage{graphicx}
\usepackage{cite}
\usepackage{enumitem}
\usepackage{hyperref}
\usepackage{float}
\usepackage{adjustbox}
\usepackage{placeins}
\usepackage{booktabs}
\usepackage{tabularx}
\usepackage{threeparttable}
\usepackage[table]{xcolor}
\usepackage{xcolor}
\usepackage{multirow}

\definecolor{TableNavy}{HTML}{17365D}
\definecolor{SectionBlue}{HTML}{DCE6F1}
\definecolor{AlternateRow}{HTML}{F7FAFD}
\definecolor{OursRow}{HTML}{E6F1FA}
\definecolor{GainGreen}{HTML}{18794E}
\definecolor{DraftRed}{HTML}{A33A3A}

\definecolor{AblationBlue}{HTML}{E8F1F8} \definecolor{PlaceholderGray}{HTML}{686868}

\definecolor{AblationBlue}{HTML}{E8F1F8} \definecolor{PlaceholderGray}{HTML}{686868} 
\newcommand{\abdelta}[1]{%
\,{\tiny\textcolor{PlaceholderGray}{\textbf{#1}}}%
}

\newcommand{\convfig}{%
    \raisebox{-0.3\height}{%
        \includegraphics[height=1.5em]{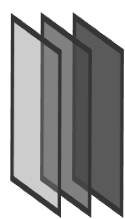}%
    }%
}

\newcommand{\elementadd}{%
    \raisebox{-0.3\height}{%
        \includegraphics[height=1.5em]{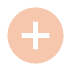}%
    }%
}

\newcommand{\convbound}{%
    \raisebox{-0.3\height}{%
        \includegraphics[height=1.5em]{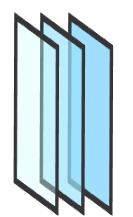}%
    }%
}

\newcommand{\boundfused}{%
    \raisebox{-0.3\height}{%
        \includegraphics[height=1.5em]{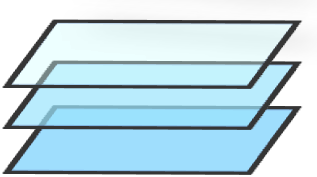}%
    }%
}

\newcommand{\concat}{%
    \raisebox{-0.3\height}{%
        \includegraphics[height=1.5em]{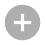}%
    }%
}

\newcommand{\mult}{%
    \raisebox{-0.3\height}{%
        \includegraphics[height=1.5em]{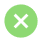}%
    }%
}

\newcommand{\reshead}{%
    \raisebox{-0.3\height}{%
        \includegraphics[height=1.5em]{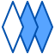}%
    }%
}

\newcommand{\gain}[1]{%
    \,{\scriptsize\textcolor{DeltaGray}{%
    (\textbf{#1$\uparrow$})}}%
}

\newcolumntype{Y}{>{\centering\arraybackslash}X}

\definecolor{OursBlue}{HTML}{E8F1F8}
\definecolor{DeltaGray}{HTML}{666666}
\definecolor{DraftGray}{HTML}{888888}

\def\tsc#1{\csdef{#1}{\textsc{\lowercase{#1}}\xspace}}
\tsc{WGM}
\tsc{QE}

\begin{document}

\title{\textbf{DERA}: Detached Edge-Residual Adaptation for\\ Prohibited item Detection}


\author{Yonathan Michael\orcidlink{0009-0005-8457-294X}, Mohamad Alansari\orcidlink{0000-0003-2960-2972}, Mohammed Bennamoun\orcidlink{0000-0002-6603-3257},~\IEEEmembership{Senior Member,~IEEE,} Dwarikanath Mahapatra\orcidlink{0000-0001-9749-7858}, Andreas Henschel\orcidlink{0000-0003-1386-5372}, Naoufel Werghi\orcidlink{0000-0002-5542-448X},~\IEEEmembership{Senior Member,~IEEE}\\ \href{https://yonathan-kiflom.github.io/DERA/page/}{\textcolor{orange}{Project Page}}}

\markboth{}%
{Anonymous Submission: DERA}



\maketitle

\begin{abstract}
Prohibited-item detection in X-ray imagery remains challenging due to object superposition, weak texture, and material clutter which obscure both semantic appearance and object boundaries. We propose \textbf{DERA}, a \textbf{D}etached \textbf{E}dge-\textbf{R}esidual \textbf{A}daptation framework for prohibited item detection under X-ray imagery. DERA combines hierarchical visual features with a parallel pixel-difference edge pyramid and learns an object-specific boundary prior from training-time contours of the instance masks. The detached prior gates edge-sensitive features, which are injected into the early visual stages through residual heads. This staged design preserves the foundation detector at the start of adaptation, isolates boundary supervision from semantic feature learning, and restricts the final adaptation stage to only \(14.7\)K trainable
parameters. Evaluated on PIDray, CLCXray, and STCray, DERA improves the baseline by \textbf{3.1}, \textbf{1.6}, and \textbf{2.4} AP points, respectively. 
\end{abstract}


\section{Introduction}
\label{intro}

\IEEEPARstart{A}{utomated} prohibited-item detection in X-ray baggage imagery is essential for transportation and public security, yet it remains substantially more difficult than conventional object detection \cite{akcaysurvey}. Because transmission images superimpose the contents of a bag, prohibited items may exhibit weak texture, transparent appearance, blurred boundaries, and severe interference from surrounding materials. Large-scale studies have identified class imbalance, background clutter, and object overlap as fundamental challenges in operational screening data \cite{sixray}, while occlusion-oriented benchmarks have shown that localization performance deteriorates as concealment becomes more severe \cite{opixray}. Subsequent datasets have moved the field toward high-quality real-world detection, deliberately hidden threats, and instance-level annotations \cite{hixray,pidray}. Nevertheless, small and thin items remain particularly difficult, with benchmark analyses showing a strong association between missed detections and reduced object size \cite{isaacmedina2023seeing}.

Research in this field has progressed from convolutional feature refinement and occlusion-aware attention toward domain-specific fusion, transformer detection, and multimodal reasoning. Class-balanced refinement and attention mechanisms have been used to suppress irrelevant background responses and recover discriminative regions under overlap \cite{sixray,opixray,clcxray}. Edge-material fusion and X-ray-specific image synthesis have further introduced structural and imaging priors into the detection pipeline \cite{jing2023emyolo,duan2023rwscfusion}. In parallel, end-to-end transformer detectors have reformulated object detection as direct set prediction \cite{detr,zhang2023dino}, while hierarchical vision transformers have strengthened multi-scale representation learning \cite{liu2021swin}. Recent X-ray adaptations explicitly address feature coupling and inaccurate localization caused by overlapping objects \cite{li2025aodetr}. Vision-language detectors additionally enable category-prompt-conditioned localization \cite{liu2024groundingdino,regionclip}, and emerging studies are extending security screening toward open-vocabulary recognition \cite{garciafernandez2025raxo}, self-supervised adaptation \cite{xssl,xthreatdet,xsslext}, dual-view detection \cite{tao2025dualview}, and cross-view and multimodal reasoning \cite{peng2026secondview,falcon}.

\begin{figure}[!t]
\centering
\includegraphics[width=0.99\linewidth]{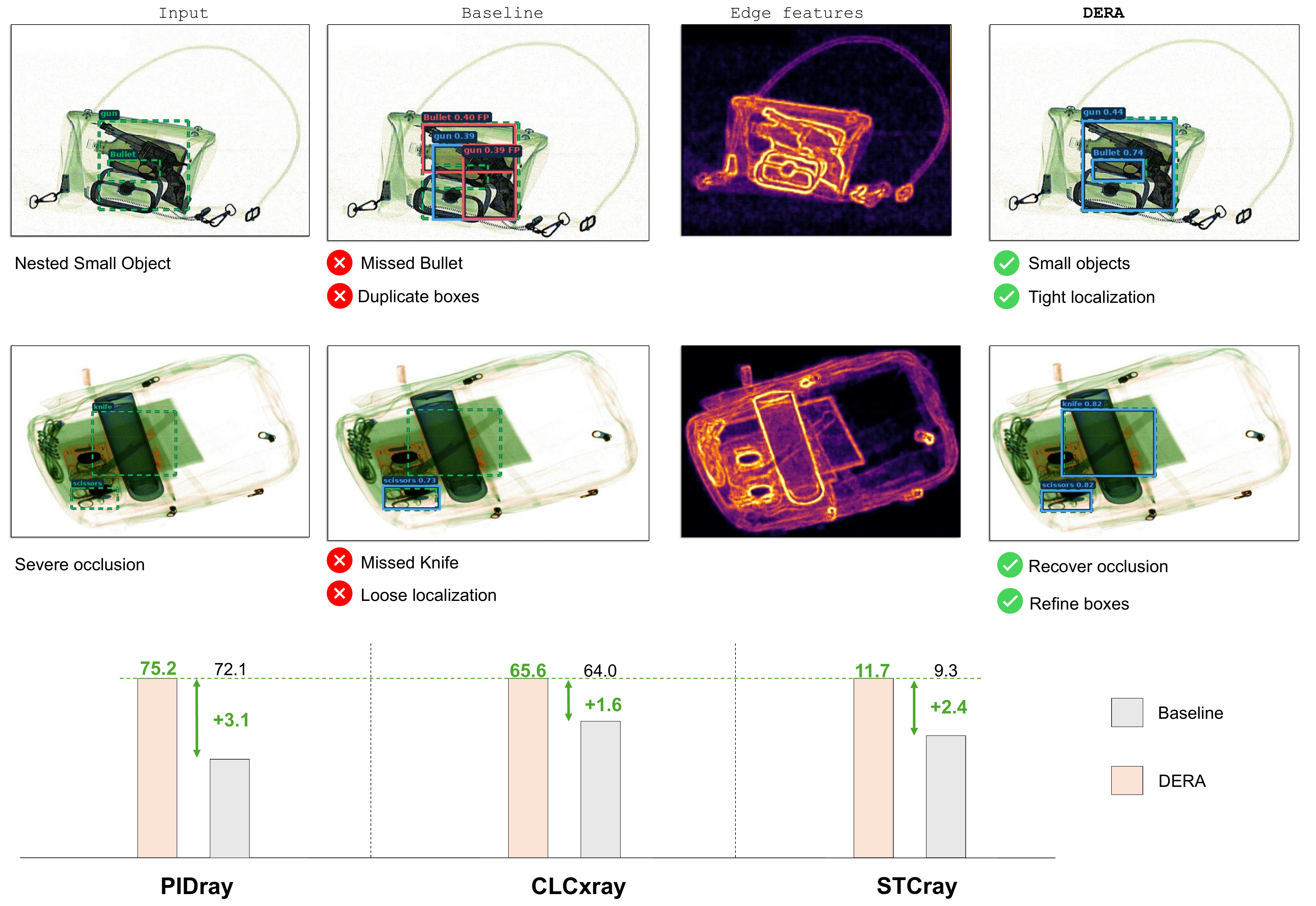}
\vspace{-1.5em}
\caption{\textbf{Motivation for DERA.} Edge-guided adaptation recovers missed and occluded prohibited items and improves localization over the baseline. Green denotes ground truth, red incorrect predictions, and blue correct predictions. Bottom panel shows AP gains across the three datasets.}
\label{fig:intro}
\vspace{-1em}
\end{figure}

Despite this progress, semantic and contextual representations may still provide imprecise spatial support when threat boundaries are fragmented by clutter or transmission overlap. Dedicated shape streams and pixel-difference operators demonstrate that explicit boundary information can complement high-level visual features \cite{takikawa2019gatedscnn,su2021pidinet}, and recent X-ray detectors similarly report benefits from edge-enhanced multi-scale processing \cite{jing2023emyolo,xu2025empdnet,mhtx,taimursupervised}. However, X-ray images also contain strong contours from benign objects, luggage structures, electronic components, and material transitions. Consequently, indiscriminate edge fusion can amplify irrelevant responses. The problem becomes more pronounced when boundary supervision is derived from imperfect pseudo-masks and jointly propagated through an already trained detector. Related multi-task studies show that competing objectives can introduce negative transfer and conflicting gradients \cite{standley2020tasks,yu2020pcgrad}, whereas side-network and adapter-based approaches suggest that lightweight residual pathways can adapt frozen representations with less destructive modification \cite{zhang2020sidetuning,chen2022adaptformer}. These observations motivate learning boundary information separately and allowing it to influence detection only through a restricted, detection-optimized residual correction.

To this end, we propose \textbf{DERA}, a
\textbf{D}etached \textbf{E}dge-\textbf{R}esidual
\textbf{A}daptation framework for prohibited-item detection in X-ray imagery. Lightweight boundary heads learn object contours from inner boundaries of the mask instances, but the boundary objective is prevented from modifying the established detector through feature detachment and parameter freezing. The learned contour prior gates detached edge-sensitive features before zero-initialized residual adapters inject a correction into the early contextual stages (Fig. \ref{fig:intro}). Thus, the enhanced architecture exactly preserves the foundation detector at the start of adapter optimization, while subsequent adapter-only training retains only boundary-guided corrections that improve the detection objective. The main contributions are:
\begin{itemize}
    \item We develop a dual-stream prohibited item detector that combines hierarchical contextual modeling with a multi-scale pixel-difference feature pyramid for complementary semantic and geometric representation.

    \item We introduce a detached boundary-guided residual mechanism in which instance-mask-derived contours gate edge features and affect the detector only through zero-initialized lightweight adapters.

    \item We propose a controlled multi-stage training strategy that isolates foundation adaptation, boundary learning, and residual tuning.
\end{itemize}

\section{Related Work}
\label{related}

\subsection{X-Ray Prohibited-Item Detection}
\label{subsec:xray_detection}

\noindent Research on prohibited-item detection has evolved alongside increasingly
realistic X-ray benchmarks \cite{divyasurvey}. Early large-scale studies established class
imbalance, object superposition, and background clutter as defining
characteristics of operational screening imagery
\cite{akcaysurvey,sixray}. Subsequent benchmarks placed greater
emphasis on severe occlusion and deliberately concealed threats
\cite{opixray}, while high-quality airport imagery and instance-level
annotations enabled more rigorous evaluation of localization in realistic
baggage scenes \cite{hixray,pidray}. Statistical analyses of
these datasets further indicate that small object size is strongly associated
with missed detections, confirming that coarse semantic recognition alone is
insufficient for reliable screening \cite{isaacmedina2023seeing}. These
benchmark developments encouraged a corresponding progression in model design where
early convolutional systems focused on class-balanced feature refinement and
spatial attention, whereas later methods introduced inhibition mechanisms,
dense multi-scale aggregation, and X-ray-specific data synthesis to improve
feature discrimination under overlap
\cite{opixray,hixray,duan2023rwscfusion}. Taken together, this
line of research shifted the problem from image-level threat recognition
toward precise object localization under clutter, concealment, and limited
foreground evidence.

Transformer-based detection subsequently changed the emphasis from local
feature enhancement to global relation modeling and end-to-end set prediction.
DETR \cite{detr} removed hand-designed proposal and suppression stages
, while Deformable DETR improved multi-scale spatial
sampling and small-object detection \cite{ddetr}. DINO further
strengthened query initialization, denoising training, and iterative box
refinement \cite{zhang2023dino}, and hierarchical backbones such as Swin
Transformer made transformer representations more suitable for dense,
multi-resolution prediction \cite{liu2021swin}. These developments have begun
to influence X-ray inspection, where transformer detectors have been adapted
to reduce foreground-background feature coupling and improve localization
under severe overlap \cite{li2025aodetr}. In parallel, contrastive
vision-language pre-training introduced transferable semantic representations
\cite{radford2021clip}, and grounded detection frameworks connected textual
phrases with localized image regions \cite{kamath2021mdetr,li2022glip}.
Grounding DINO combines this language grounding capability with DINO-style
object queries, providing a flexible foundation for category-prompt-conditioned
detection \cite{liu2024groundingdino}.

\noindent\textit{Motivation gap.}
Although these approaches improve semantic discrimination, overlap handling,
and multimodal reasoning, they leave open how explicit object-boundary evidence
can be incorporated into a single-view text-conditioned detector when the
spatial support of a prohibited item is weak or fragmented.

\subsection{Edge-Aware Representation and Controlled Adaptation}
\label{subsec:edge_adaptation}

\noindent Explicit boundary modeling has long been used to complement semantic
representations in dense visual prediction. Deeply supervised edge learning
first demonstrated that hierarchical side outputs could recover boundaries at
multiple abstraction levels \cite{xie2015hed}. Later approaches enriched this
formulation by aggregating features across convolutional depths
\cite{liu2017rcf} and assigning scale-specific supervision to different
network stages \cite{he2019bdcn}. Edge information was subsequently integrated
into higher-level tasks through dedicated shape streams, showing that semantic
and geometric processing can remain complementary rather than being compressed
into a single representation \cite{takikawa2019gatedscnn}. PiDiNet advanced
this direction by embedding classical gradient-sensitive operations into
lightweight pixel-difference convolutions, providing efficient local transition
modeling with a small parameter footprint \cite{su2021pidinet}. Parallel
body-edge representations have also been investigated in segmentation, where
hierarchical transformer context is combined with pixel-difference features
\cite{manzari2024befunet}, while \cite{pidnet} separates contextual, detail, and boundary representations and uses boundary attention to regulate their fusion for real-time semantic
segmentation.

The value of edge information is particularly relevant to X-ray imagery,
where object texture is weak but portions of a threat contour may remain
visible. Boundary activation has therefore been used to improve feature
discrimination in cluttered security images \cite{hixray,mhtx}, while other
detectors combine edge and material representations to strengthen recognition
under overlap \cite{jing2023emyolo}. Recent multi-scale architectures similarly
introduce Sobel-based or residual edge branches to recover spatial detail
suppressed by deep feature extraction \cite{xu2025empdnet}. Strong X-ray edges often arise from benign structures and material transitions rather than threat boundaries; therefore, directly fusing generic edge cues with semantic features may amplify clutter, particularly under joint optimization.

This challenge parallels multi-task learning, where auxiliary-task benefits depend on task relatedness, balanced loss scales, and compatible gradients. Gradient normalization
methods address unequal learning rates across objectives
\cite{chen2018gradnorm}, while empirical task-grouping studies show that
apparently related visual tasks do not always benefit from unrestricted
parameter sharing \cite{standley2020tasks}. Multi-task optimization studies show that conflicting gradients from
different objectives can cause negative transfer when they jointly update
shared parameters \cite{yu2020pcgrad}. A complementary research direction avoids
such interference by keeping the pre-trained representation fixed \cite{liu2021cagrad}.
Residual adapters introduce domain-specific corrections through small additive
modules \cite{rebuffi2017adapters}, and side-network adaptation separates new
task learning from the original network pathway
\cite{zhang2020sidetuning}. Visual prompt tuning and transformer adapters
similarly adapt large visual models while freezing most or all of the
pre-trained backbone \cite{jia2022vpt,chen2022adaptformer}. More recent
parallel adaptation architectures additionally avoid back-propagating through
the frozen backbone, improving parameter, memory, and training-time efficiency
\cite{mercea2024visualadaptation}.

\noindent\textit{Motivation gap.}
Existing edge-enhanced X-ray detectors generally couple structural feature
learning with detector optimization, whereas adaptation research rarely
considers uncertain object-boundary supervision. This motivates learning
instance mask-derived contours in an isolated branch and transferring their
influence through detached, zero-initialized residual adapters.

\section{\textbf{DERA}: \textbf{D}etached \textbf{E}dge-\textbf{R}esidual
\textbf{A}daptation}
\label{sec:method}

\subsection{Overview}
\label{sec:method_overview}

\begin{figure*}[!t]
\centering
\includegraphics[width=0.99\linewidth,height=7.5cm]{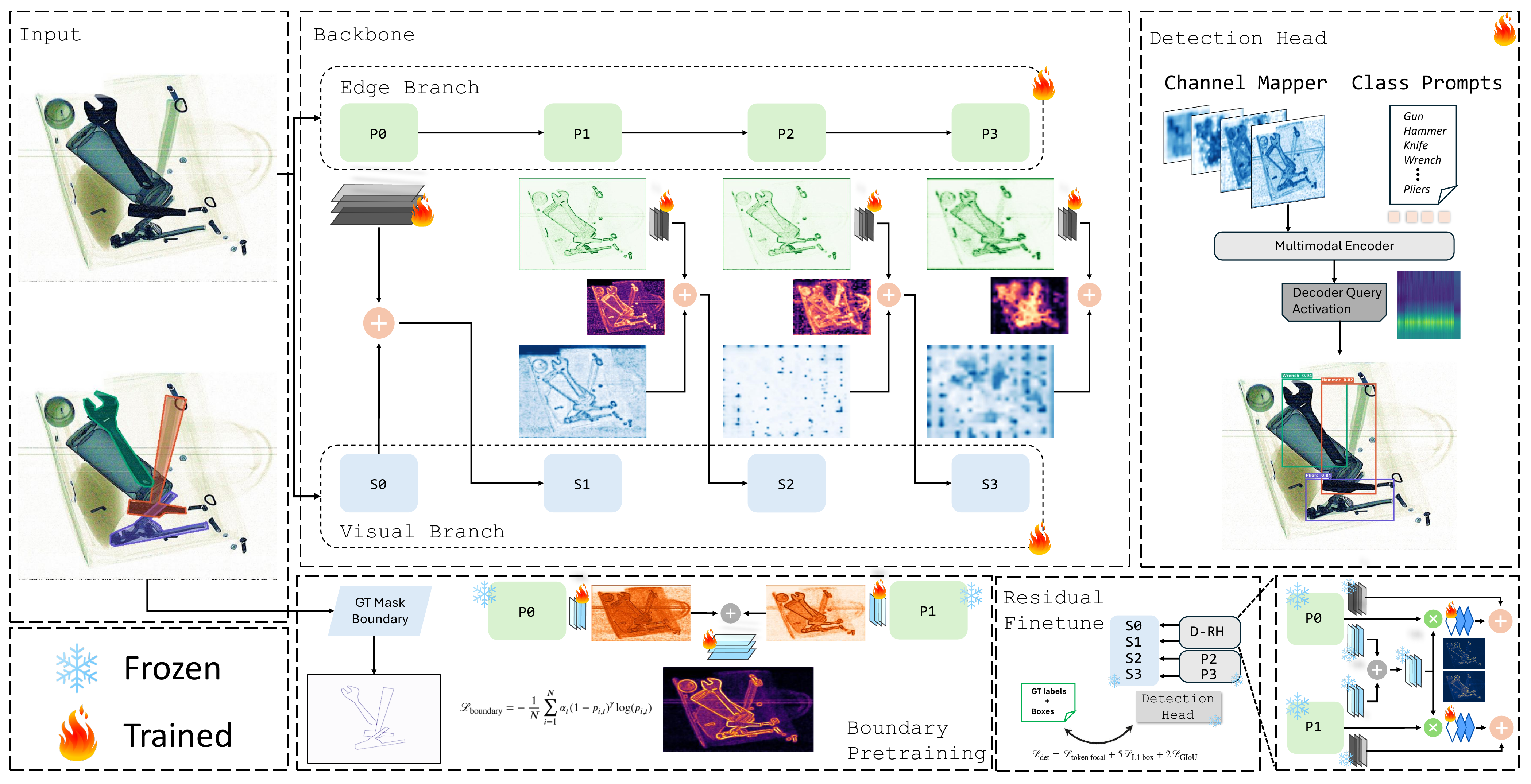}
\vspace{-0.5em}

\caption{\textbf{Overview of DERA.} Visual and edge-branch features are fused for prompt-conditioned detection, while a contour-guided boundary prior gates residual corrections injected only into the early visual stages.}
\label{fig:block}
\vspace{-1em}
\end{figure*}

\noindent Given an X-ray image \(I\) and category prompts
\(\mathcal{T}=\{t_c\}_{c=1}^{C}\), our framework predicts prompt-aligned class scores and bounding boxes. As shown in Fig. \ref{fig:block}, the model contains two parallel branches: a visual branch of hierarchical transformers that captures contextual information and
an edge branch of pixel-difference network that emphasizes local intensity transitions.
Their multiscale features are fused and processed by a text-conditioned
object detector. To distinguish object contours from generic X-ray edges, we introduce a
boundary-prior branch supervised by contours derived from instance-masks. The predicted boundary probability gates early
pixel-difference features, which are injected into the detector through
zero-initialized residual projections. Instance-masks are used only to extract target contours during
training and are not required at inference.

\subsection{Dual-Stream Foundation Detector}
\label{sec:foundation_detector}

\subsubsection{Contextual and pixel-difference features}

\noindent Given an X-ray image, the visual branch produces feature maps \(\{S_i\}_{i=0}^{3}\), while the parallel pixel-difference branch produces the corresponding edge-sensitive pyramid \(\{P_i\}_{i=0}^{3}\). For an input feature map \(X\), each pixel-difference block is formulated as:
\begin{equation} Y = \mathcal{R}(X) + \operatorname{Conv}_{1\times1} \left[ \operatorname{ReLU} \left( \mathcal{K}_{3\times3}(X) \right) \right], \label{eq:pdc_block} \end{equation}
where \(\mathcal{K}_{3\times3}\) denotes a depthwise central, angular, or radial difference operator, or an ordinary convolution according to the predefined block sequence. \(\mathcal{R}(\cdot)\) is an identity
mapping when the spatial resolution is preserved; in downsampling blocks, it
applies max pooling followed by a \(1\times1\) projection. The outputs collected at the four hierarchy levels form the
edge-sensitive pyramid \(\{P_i\}_{i=0}^{3}\). Alternating pixel-difference and ordinary convolutions allows the branch to retain both local transition cues and conventional appearance information.

\subsubsection{Cross-Stream Fusion}

\noindent At each hierarchy level, \(P_i\) is aligned with \(S_i\) using a a stage-specific \(1\times1\)
projection \convfig \space and optional bilinear resizing. The two streams are then fused by elementwise addition \elementadd as: \begin{equation} B_i = S_i \oplus \mathcal{A}_i(P_i), \qquad i\in\{0,1,2,3\}, \label{eq:foundation_fusion} \end{equation} where \(\mathcal{A}_i(\cdot)\) denotes the projection and spatial alignment operation. The fused pyramid \(\{B_i\}_{i=0}^{3}\) forms the multiscale visual representation used by the subsequent detection head.

\subsubsection{Detection Head}

\noindent The detection head follows the baseline \cite{liu2024groundingdino} and remains unchanged. The fused
pyramid \(\{B_i\}_{i=0}^{3}\) is projected to a shared embedding space and
jointly encoded with the category prompts. An object-query decoder then
produces \(N\) query embeddings \(\{z_q\}_{q=1}^{N}\). For each query, the
classification scores and bounding box are predicted as:
\begin{equation}
    \mathbf{s}_q = \mathcal{C}(z_q,T),
    \qquad
    \hat{\mathbf{b}}_q = \mathcal{B}(z_q),
    \label{eq:detection_head}
\end{equation}
where \(T\) denotes the encoded prompt tokens,
\(\mathcal{C}(\cdot)\) is the text-conditioned classification branch, and
\(\mathcal{B}(\cdot)\) predicts the normalized box coordinates
\((c_x,c_y,w,h)\). The token-level scores are mapped to their corresponding
category prompts to obtain the final detections. The detector is optimized using the token-level focal classification loss
together with \(\ell_1\) and generalized IoU losses for box regression:
\begin{equation}
    \mathcal{L}_{\mathrm{det}}
    =
    \mathcal{L}_{\mathrm{focal}}^{\mathrm{tok}}
    +
    5\mathcal{L}_{\ell_1}^{\mathrm{box}}
    +
    2\mathcal{L}_{\mathrm{GIoU}}.
    \label{eq:detection_loss}
\end{equation}

\subsubsection{Boundary-Prior Prediction}
\noindent The boundary branch uses the two highest-resolution pixel-difference features,
\(P_0\) and \(P_1\). After stop-gradient and spatial alignment, each feature
is processed by an independent side head \convbound:
\begin{equation}
    L_i =
    \mathcal{H}_i
    \left(
        \operatorname{Resize}
        \left(
            \operatorname{sg}(P_i)
        \right)
    \right),
    \qquad i\in\{0,1\},
    \label{eq:boundary_side_head}
\end{equation}
where \(\mathcal{H}_i\) is a \(3\times3\) convolution, group
normalization, ReLU, and a \(1\times1\) output projection. The aligned side logits are concatenated \concat and fused using a learned \(1\times1\) convolution \boundfused: 
\begin{equation} L = \mathcal{F}\!\left(L_0 \mathbin{\|} L_1\right), \qquad A = \operatorname{sg}\!\left(\sigma(L)\right), \label{eq:boundary_prior} \end{equation} 
where \(\mathbin{\|}\) denotes channel-wise concatenation, \(\mathcal{F}\) is the fusion projection, and \(A\) is the detached boundary prior. The fusion weights are initialized to combine the two scales equally. Detaching \(P_i\) prevents the
boundary loss from modifying the feature extractor, while detaching \(A\)
prevents subsequent detection optimization from updating the boundary branch.

\subsubsection{Edge Gating and Residual Injection}

\noindent At the first two hierarchy levels, the detached boundary prior \(A\) is resized
and broadcast across the channels of the corresponding pixel-difference
feature. The gated feature, residual correction, and final injection are
defined as:
\begin{equation}
\begin{aligned}
    G_i &= \operatorname{sg}(P_i)\odot\rho_i(A), \\
    R_i &= \mathcal{Z}_i(G_i), \\
    X_i &= B_i\oplus R_i,
\end{aligned}
\qquad i\in\{0,1\},
\label{eq:edge_residual_injection}
\end{equation}
where \(\odot\) denotes element-wise multiplication \mult,
\(\oplus\) denotes element-wise addition \elementadd, and \(\rho_i(\cdot)\) resizes the
boundary prior to the resolution of \(P_i\). Each
\(\mathcal{Z}_i(\cdot)\) \reshead is a stage-specific residual head implemented as a
\(1\times1\) projection to the channel dimension of \(B_i\). Its weights and
biases are initialized to zero, yielding \(R_i=0\) and \(X_i=B_i\) before
residual training. The subsequent optimization therefore updates only the
residual heads to learn boundary-guided corrections to the foundation
features.

\vspace{-1em}
\subsection{Boundary Supervision}
\label{sec:boundary_supervision}

\begin{figure}[t]
\vspace{-0.5em}
\centering
\includegraphics[width=0.99\linewidth]{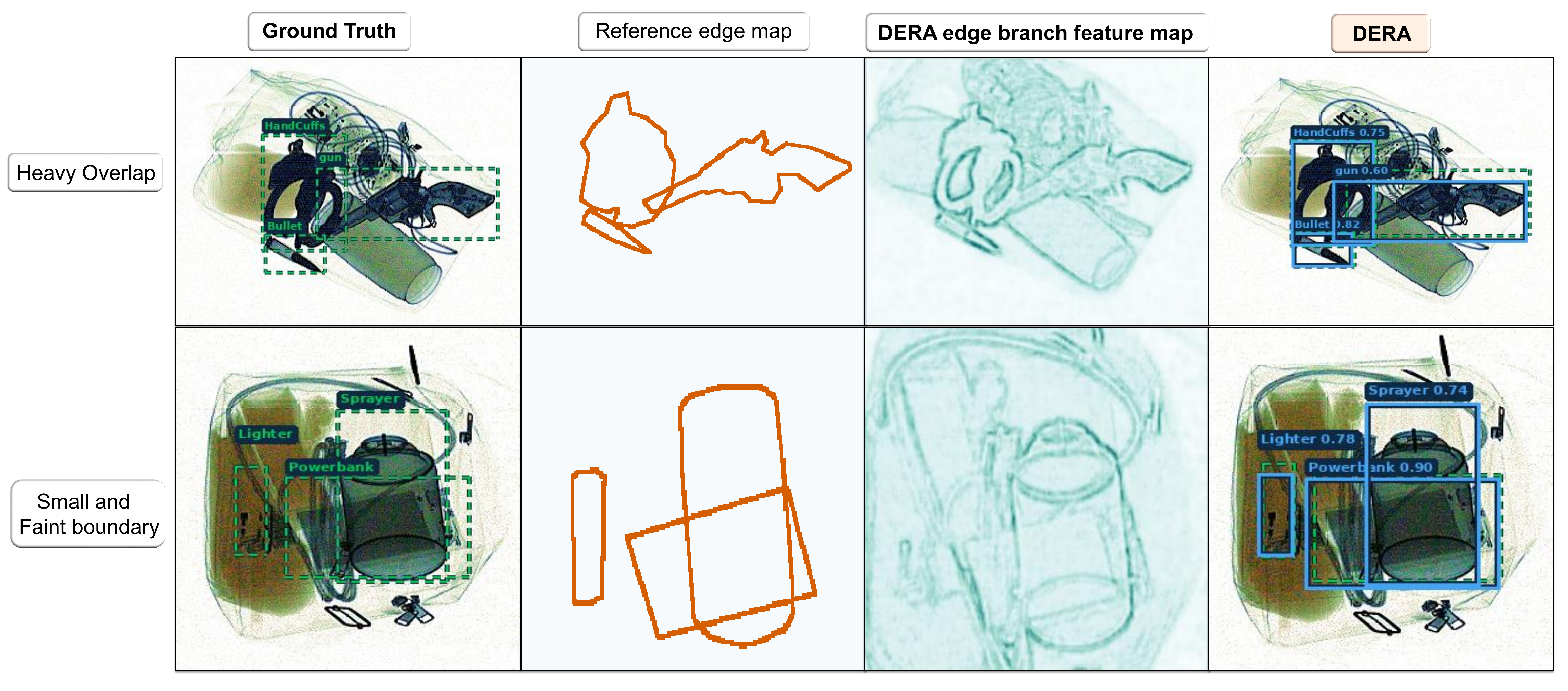}
\vspace{-2em}
\caption{\textbf{Boundary-guided feature visualization.} DERA’s edge features preserve object structure under heavy overlap and faint boundaries, supporting accurate detections. Green dashed and blue boxes denote ground truth and DERA predictions, respectively.}
\label{fig:edgemaps}
\vspace{-1em}
\end{figure}

\noindent For each object, a thin inner contour is extracted by subtracting the inner pixels of the original mask (Fig. \ref{fig:edgemaps}). The instance contours are
merged, resized to the boundary-logit resolution using adaptive max pooling,
and masked to exclude padded image regions. The boundary branch is optimized using focal loss:
\begin{equation}
    \mathcal{L}_{\mathrm{boundary}}
    =
    -\frac{1}{N}
    \sum_{i=1}^{N}
    \alpha_t
    \left(1-p_{i,t}\right)^{\gamma}
    \log\left(p_{i,t}\right),
    \label{eq:boundary_loss}
\end{equation}
where \(p_{i,t}\) is the predicted probability assigned to the target class
at pixel \(i\), \(\alpha_t\) is its class-balancing factor, and \(N\) is the
number of valid pixels contributing to the loss. We use
\(\alpha=0.75\) and \(\gamma=2\), while padded regions are excluded from
optimization. The contours derived from instance masks are required only for boundary pretraining
and are not used during inference.

\subsection{Training Strategy}
\label{sec:training_strategy}

\noindent The model is optimized in three consecutive stages to separate semantic
detector adaptation from auxiliary boundary learning.

\subsubsection{Foundation Adaptation}

\noindent The visual and edge branches, cross-stream projections, and
detection modules are jointly adapted to the X-ray domain, while the language
encoder remains frozen. A lower learning rate is applied to the visual
backbone. Training loss uses the detection objective detailed in Eq. \ref{eq:detection_loss}.

\subsubsection{Boundary-Head Learning}

\noindent The foundation checkpoint initializes the boundary branch. All foundation
components and residual heads are frozen, while only the side heads
\(\{\mathcal{H}_i\}_{i=0}^{1}\) and fusion layer \(\mathcal{F}\) are
optimized using \(\mathcal{L}_{\mathrm{boundary}}\) detailed in Eq. \ref{eq:boundary_loss}. This stage learns the
boundary prior without allowing mask-derived supervision to modify the
established detector.

\subsubsection{Residual Adaptation}

\noindent The trained boundary branch is subsequently frozen together with the complete
foundation model. Only the zero-initialized residual heads
\(\{\mathcal{Z}_i\}_{i=0}^{1}\) are optimized using
\(\mathcal{L}_{\mathrm{det}}\) in Eq. \ref{eq:detection_loss}. Detection gradients propagate through the
frozen network to the residual injection points, but update only the residual
heads. The final stage therefore learns boundary-guided corrections according
to their contribution to classification and box localization.

\subsection{Inference}
\label{sec:inference}

\noindent Inference requires only an X-ray image and the category prompts. The
visual and edge branches first generate the fused visual
pyramid \(\{B_i\}_{i=0}^{3}\), while the frozen boundary branch predicts the
internal prior \(A\) from \(P_0\) and \(P_1\). The edge-residual pathway
operates only at the first two visual hierarchy stages: \(A\) gates the
corresponding edge features, and the trained residual heads inject
the resulting corrections into \(B_0\) and \(B_1\) through element-wise
addition. The deeper features \(B_2\) and \(B_3\) remain unchanged, yielding
the final pyramid \(\{X_0,X_1,B_2,B_3\}\). This representation is passed with
the category prompts to the detection head, which produces prompt-aligned
class scores and bounding boxes. No instance masks, contour targets, or
external edge maps are required at inference.

\section{Experiments}
\label{experiments}

\subsection{Implementation Details}
\label{implementation}

\noindent We implement the proposed framework on Grounding DINO
\cite{liu2024groundingdino}, using a Swin Transformer
\cite{liu2021swin} as the hierarchical contextual backbone and
adapting PiDiNet \cite{su2021pidinet} as the parallel pixel-difference edge
pyramid. Training follows three sequential stages. During foundation
adaptation, the PiDiNet stream, cross-stream projections, Swin backbone,
visual channel mapper, text-feature projection, multimodal encoder, query
decoder, and detection heads are optimized using token-level focal
classification, \(L_1\), and GIoU losses weighted by \(1\), \(5\), and
\(2\), respectively. The Swin backbone is updated with a lower learning
rate than the newly introduced modules, while the language backbone remains
frozen. The resulting checkpoint initializes boundary-head learning, where
only the two boundary side heads and their \(1\times1\) logit-fusion layer
are trained using sigmoid focal loss with \(\alpha=0.75\) and
\(\gamma=2\); the Grounding DINO foundation and PiDiNet feature stream
remain fixed. Finally, the learned boundary head is frozen and only the
zero-initialized residual \(1\times1\) projections are optimized using the
original Grounding DINO detection objective. Checkpoints are transferred
sequentially between stages, and the residual-adapted checkpoint with the
best validation performance is used for evaluation.

\subsection{Experimental Setup}
\label{experimental_setup}

\noindent All experiments were conducted on a
workstation equipped with a single NVIDIA GeForce RTX 3090 GPU with
24\,GB of VRAM. Training followed the three-stage procedure described in
Section~\ref{sec:training_strategy}, with checkpoints transferred
sequentially between stages. During foundation adaptation, the detector and
pixel-difference stream are optimized using token-level focal
classification, \(L_1\), and GIoU losses weighted by \(1\), \(5\), and
\(2\), respectively, with learning rates of
\texttt{$1 \times 10^{-4}$} for newly introduced modules and
\texttt{$1 \times 10^{-5}$} for the Swin backbone. Boundary-head learning uses
sigmoid focal loss with \(\alpha=0.75\) and \(\gamma=2\), a learning rate
of \texttt{$1 \times 10^{-4}$}, and updates only the boundary side heads and
their fusion layer. Edge-gated residual adaptation subsequently freezes the
foundation detector and boundary head and optimizes only the
zero-initialized residual projections using a learning rate of
\texttt{$1 \times 10^{-4}$} and the original detection objective. The three
stages are trained for 15,
2, and 3 epochs,
respectively, using a batch size of 4 and
AdamW optimization. Peak GPU-memory usage is
$\sim10\,GB$,
$\sim3\,GB$, and
$\sim6\,GB$ for the three training stages,
respectively, and $\sim4\,GB$ during inference.
The residual-adapted checkpoint with the best validation performance is used
for final evaluation. Instance masks are required only to extract the contour maps for boundary-head supervision during training; inference uses only an X-ray
image and category prompts.

\begin{table}[ht]
\centering
\caption{Computational Cost and Adaptation Efficiency of training and Inference using setup detailed in Sec. \ref{experimental_setup}}
\label{tab:computational_efficiency}

\scriptsize
\setlength{\tabcolsep}{2.0pt}
\renewcommand{\arraystretch}{1.15}

\begin{threeparttable}

\begin{adjustbox}{max width=\columnwidth}
\begin{tabular}{@{}c l c r rr rrrr@{}}

\toprule

&
\multirow{2}{*}{\textbf{Stage / Model}}
&
\multirow{2}{*}{\textbf{Prec.}}
&
\multirow{2}{*}{\textbf{Params (M)}}
&
\multicolumn{2}{c}{\textbf{Compute}}
&
\multicolumn{4}{c}{\textbf{Runtime}}
\\

\cmidrule(lr){5-6}
\cmidrule(lr){7-10}

&
&
&
&
\textbf{GMAC/img}
&
\textbf{GFLOP/img}
&
\textbf{ms/batch}
&
\textbf{ms/img}
&
\textbf{img/s}
&
\textbf{Peak memory}
\\

\midrule

\multirow{4}{*}{
    \rotatebox[origin=c]{90}{\textbf{Training}}
}
&
Grounding DINO
&
FP32
&
36.503
&
329.57
&
659.14
&
1260.75
&
315.19
&
3.17
&
7.06
\\

&
BEF-Swin
&
AMP
&
64.336
&
349.34
&
698.68
&
1009.94
&
252.49
&
3.96
&
9.51
\\

&
Boundary pretraining
&
FP32
&
0.026
&
13.17
&
26.34
&
63.69
&
15.92
&
62.81
&
2.13
\\

&
Residual fine-tuning
&
FP32
&
\textbf{0.015}
&
350.97
&
701.94
&
1041.03
&
260.26
&
3.84
&
5.11
\\

\midrule

\multirow{2}{*}{
    \rotatebox[origin=c]{90}{\textbf{Inf.}}
}
&
Grounding DINO
&
FP32
&
172.915
&
436.69
&
873.37
&
421.92
&
105.48
&
9.48
&
2.78
\\

&
\textbf{DERA}
&
FP32
&
\textbf{173.269}
&
\textbf{465.55}
&
\textbf{931.11}
&
\textbf{512.42}
&
\textbf{128.10}
&
\textbf{7.81}
&
\textbf{3.37}
\\

\bottomrule

\end{tabular}
\end{adjustbox}
\end{threeparttable}
\vspace{-1.5em}
\end{table}

\subsection{Computational Efficiency}
\label{subsec:efficiency}

Table~\ref{tab:computational_efficiency} details the breakdown of computational cost and adaptation efficiency of each stage in DERA. Boundary pretraining and residual
fine-tuning update only \(0.026\)M and \(0.015\)M parameters, respectively.
Boundary pretraining is also computationally inexpensive as it executes
only the isolated boundary pathway. Residual fine-tuning, however, still
evaluates the complete frozen detector so that the residual heads can be
optimized using the detection objective and its forward cost therefore remains
comparable to foundation adaptation despite the small trainable parameter
count. At inference, DERA adds only \(0.354\)M parameters (\(0.20\%\)) and
\(28.86\) GMAC/img (\(6.6\%\)) over the baseline. This increases amortized
latency from \(105.48\) to \(128.10\) ms/img and peak memory from \(2.78\) to
\(3.37\) GB.

\vspace{-1em}

\subsection{Datasets}
\label{sec:datasets}

\noindent \textbf{PIDray} \cite{pidray} is a large-scale real-world X-ray benchmark
containing 47,677 images from 12 prohibited-item categories, with
instance-level bounding-box and segmentation-mask annotations. The dataset
comprises 29,457 training images and 18,220 test images, with the test set divided into Easy, Hard, and Hidden subsets according to object
clutter and concealment difficulty. It therefore provides a challenging
evaluation of detection under occlusion, class imbalance, and deliberate
threat concealment.

\noindent \textbf{CLCXray} \cite{clcxray} contains 9,565 X-ray images, including 4,543
scans collected from real subway-security scenes and 5,022 scans generated
from manually arranged baggage. The dataset covers 12 categories comprising
five types of cutters and seven types of liquid containers, with object
annotations provided in COCO bounding-box format. To align with our training objective we generate targets' pixel-level masks using a segmentation model \cite{sam2}. Its frequent overlap
between prohibited items, benign objects, and visually similar background
structures makes it particularly relevant for evaluating boundary-sensitive
detection.

\noindent \textbf{STCray} \cite{stingbee} is a multimodal X-ray security dataset
containing 46,642 image-caption pairs across 21 categories, divided
into 30,044 training and 16,598 test images. It provides bounding boxes,
pixel-level annotations, and textual descriptions, and follows the STING
collection protocol to vary threat position, orientation, clutter, and
occlusion. In our experiments, only bounding boxes and
category labels supervise detection, while pixel-level annotations are used
only to construct auxiliary boundary targets during training.

\subsection{Evaluation Metrics}
\label{subsec:metrics}

\noindent We report the COCO-style average precision metrics. AP averages precision over
IoU thresholds from \(0.50\) to \(0.95\) in increments of \(0.05\), while
AP$_{50}$ and AP$_{75}$ evaluate detection at fixed IoU thresholds of \(0.50\)
and \(0.75\), respectively. We additionally report AP$_S$, AP$_M$, and AP$_L$,
which compute AP over the same IoU range for small
(\(\mathrm{area}<32^2\)), medium
(\(32^2\leq\mathrm{area}<96^2\)), and large
(\(\mathrm{area}\geq96^2\)) objects. All results are expressed as percentages,
with higher values indicating better performance.

\section{Experimental Results and Discussion}
\label{sec:results}

\noindent The proposed method is evaluated independently on PIDray, CLCXray, and
STCray. All datasets are evaluated as class-conditioned
bounding-box detection tasks using the metrics detailed in \ref{subsec:metrics}. All reproduced methods use the same dataset partitions, image preprocessing,
and evaluator. Vision-language methods use the same category prompts.
Method-specific optimization settings follow their respective implementations.

\vspace{-0.2cm}
\begin{table*}[t]
\centering
\caption{
Evaluation of detection performance in AP, AP$_{50}$ and
AP$_{75}$ (\%). Best and second-best results are shown in bold and
underlined, respectively. Improvements in parentheses are computed relative
to the baseline Grounding DINO \cite{liu2024groundingdino}.
}
\label{tab:cross_dataset_results}

\scriptsize
\setlength{\tabcolsep}{2.8pt}
\renewcommand{\arraystretch}{1.15}

\begin{threeparttable}

\begin{adjustbox}{max width=\textwidth,center}
\begin{tabular}{@{}c l ccc ccc ccc@{}}

\toprule

&
\multirow{2}{*}{\textbf{Method}}
&
\multicolumn{3}{c}{\textbf{PIDray}}
&
\multicolumn{3}{c}{\textbf{CLCXray}}
&
\multicolumn{3}{c}{\textbf{STCray}}
\\

\cmidrule(lr){3-5}
\cmidrule(lr){6-8}
\cmidrule(lr){9-11}

&
&
\textbf{AP}
&
\textbf{AP$_{50}$}
&
\textbf{AP$_{75}$}
&
\textbf{AP}
&
\textbf{AP$_{50}$}
&
\textbf{AP$_{75}$}
&
\textbf{AP}
&
\textbf{AP$_{50}$}
&
\textbf{AP$_{75}$}

\\

\midrule

\multirow{5}{*}{
    \rotatebox[origin=c]{90}{\textbf{Vision-only}}
}
&
\mbox{Faster R-CNN~\cite{ren2015fasterrcnn}}
&
59.7
&
80.3
&
68.7
&
55.9
&
70.4
&
66.7
&
\textbf{12.7}
&
18.4
&
\underline{13.9}
\\

&
\mbox{AO-DETR~\cite{li2025aodetr}}
&
59.4
&
74.3
&
64.0
&
54.4
&
66.4
&
63.4
&
11.8
&
16.8
&
12.7
\\

&
\mbox{CSPCL~\cite{cspcl}}
&
61.6
&
75.9
&
67.0
&
55.2
&
67.9
&
63.7
&
\underline{12.3}
&
17.3
&
13.1
\\

&
\mbox{Deformable DETR~\cite{ddetr}}
&
58.3
&
67.9
&
61.1
&
51.2
&
60.6
&
57.9
&
8.2
&
11.9
&
10.3
\\

&
\mbox{DINO~\cite{zhang2023dino}}
&
59.2
&
73.1
&
62.8
&
54.3
&
67.2
&
63.6
&
11.7
&
16.4
&
12.9
\\

\midrule

\multirow{3}{*}{
    \rotatebox[origin=c]{90}{\textbf{VLM}}
}

&
\mbox{RegionClip~\cite{regionclip}}
&
28.5
&
48.7
&
27.3
&
51.0
&
65.5
&
60.4
&
5.7
&
10.7
&
5.4
\\

&
\mbox{Yolo-World~\cite{yoloworld}}
&
37.6
&
62.3
&
41.3
&
55.8
&
66.9
&
57.4
&
6.2
&
10.9
&
6.5
\\

&
\mbox{Grounding DINO~\cite{liu2024groundingdino}}
&
72.1
&
82.8
&
77.6
&
64.0
&
75.2
&
72.5
&
9.3
&
12.1
&
10.3
\\

\rowcolor{gray!15}
&
\mbox{BEF-Swin (Ours)}
&
\underline{74.5}\gain{2.4}
&
\underline{85.7}\gain{2.9}
&
\underline{79.9}\gain{2.3}
&
\underline{65.3}\gain{1.3}
&
\underline{76.9}\gain{1.7}
&
\underline{73.2}\gain{0.7}
&
11.2\gain{1.9}
&
\underline{18.8}\gain{6.7}
&
12.8\gain{2.5}
\\

\rowcolor{OursBlue}
&
\textbf{\mbox{DERA (Ours)}}
&
\textbf{75.2}\gain{3.1}
&
\textbf{86.2}\gain{3.4}
&
\textbf{80.6}\gain{3.0}
&
\textbf{65.6}\gain{1.6}
&
\textbf{77.3}\gain{2.1}
&
\textbf{74.4}\gain{1.9}
&
11.7\gain{2.4}
&
\textbf{19.5}\gain{7.4}
&
\textbf{14.5}\gain{4.2}
\\

\bottomrule

\end{tabular}
\end{adjustbox}

\end{threeparttable}
\vspace{-0.5cm}
\end{table*}


\subsection{Comparison with Existing Methods}
\label{subsec:main_comparison}

\noindent Table~\ref{tab:cross_dataset_results} shows that the proposed model provides
consistent improvements over the baseline across all three evaluation datasets. Among the vision-language models, RegionCLIP \cite{regionclip} and YOLO-World \cite{yoloworld} remain substantially below Grounding DINO, particularly on PIDray and STCray. This suggests that generic region--text alignment alone is insufficient to overcome the appearance shift, weak texture, and object superposition characteristic of security X-rays. The consistent improvement from Grounding DINO to BEF-Swin indicates that pixel-difference features supply local structural evidence that complements grounded semantic reasoning. DERA further improves every BEF-Swin result, with particularly notable AP$_{75}$ gains on CLCXray and STCray, supporting the intended role of the detached boundary prior in refining the spatial support of detected objects. On PIDray and CLCXray, DERA also outperforms the vision-only methods across all metrics, demonstrating that controlled edge adaptation can complement language-grounded features more effectively than the evaluated benchmarks.

The STCray results reveal a more nuanced behavior. FR-CNN \cite{ren2015fasterrcnn}, CSPCL \cite{cspcl}, and AO-DETR \cite{li2025aodetr} obtain slightly higher overall AP than DERA, whereas DERA achieves the strongest AP$_{50}$ and AP$_{75}$. Because COCO AP averages performance over IoU thresholds from $0.50$ to $0.95$, this ranking implies that DERA is more effective at producing detections that satisfy moderate and strict overlap criteria, but is less consistent at some of the remaining thresholds, plausibly those requiring near-exact box alignment. A likely reason is that the boundary-guided residual pathway helps recover sufficient object extent from incomplete or cluttered visual evidence, moving more predictions above the $0.50$ and $0.75$ matching thresholds, while fragmented contours, imperfect boundary supervision, or annotation ambiguity can still limit extremely high-IoU localization. The vision-only detectors may therefore retain an advantage in the final precision of box regression on STCray, whereas DERA provides stronger object recovery and practically useful localization in extreme clutter.
\vspace{-1em}
\begin{table}[ht]
\centering
\caption{
Scale-wise performance in AP$_{S}$, AP$_{M}$ and
AP$_{L}$ (\%). Best and second-best results are shown in bold and
underlined, respectively. Improvements in parentheses are computed relative
to the baseline Grounding DINO \cite{liu2024groundingdino}.}
\label{tab:scale_results}

\scriptsize
\setlength{\tabcolsep}{2.8pt}
\renewcommand{\arraystretch}{1.15}

\begin{threeparttable}
\begin{adjustbox}{max width=\columnwidth}
\begin{tabular}{@{}l ccc ccc ccc@{}}

\toprule
\multirow{2}{*}{\textbf{Method}}
&
\multicolumn{3}{c}{\textbf{PIDray}}
&
\multicolumn{3}{c}{\textbf{CLCXray}}
&
\multicolumn{3}{c}{\textbf{STCray}}
\\

\cmidrule(lr){2-4}
\cmidrule(lr){5-7}
\cmidrule(lr){8-10}

&
\textbf{AP$_S$} & \textbf{AP$_M$} & \textbf{AP$_L$}
&
\textbf{AP$_S$} & \textbf{AP$_M$} & \textbf{AP$_L$}
&
\textbf{AP$_S$} & \textbf{AP$_M$} & \textbf{AP$_L$}
\\

\midrule

Grounding DINO~\cite{liu2024groundingdino}
&
9.4 & 57.7 & 70.3
&
24.1 & 34.9 & 68.6
&
5.3 & 13.4 & 15.2
\\

\rowcolor{gray!15}
BEF-Swin (Ours)
&
\underline{12.2}\gain{2.8} & \underline{59.4}\gain{1.7} & \underline{71.9}\gain{1.6}
&
\underline{28.3}\gain{4.2} & \underline{36.8}\gain{1.9} & \underline{70.8}\gain{2.2}
&
\underline{6.5}\gain{1.2} & \underline{14.0}\gain{0.6} & \underline{16.1}\gain{0.9}
\\

\rowcolor{OursBlue}
\textbf{DERA (Ours)}
&
\textbf{14.0}\gain{4.6} & \textbf{61.6}\gain{3.9} & \textbf{73.7}\gain{3.4}
&
\textbf{31.6}\gain{7.5} & \textbf{39.4}\gain{4.5} & \textbf{71.1}\gain{2.5}
&
\textbf{7.5}\gain{2.2} & \textbf{15.1}\gain{1.7} & \textbf{16.7}\gain{1.5}
\\

\bottomrule
\end{tabular}
\end{adjustbox}
\end{threeparttable}
\vspace{-0.5cm}
\end{table}

\subsection{Performance Across Object Scales}
\label{subsec:scale_analysis}

\noindent Table~\ref{tab:scale_results} shows that DERA provides its largest gains on small objects, improving AP$_S$ over the baseline by up to \(7.5\) points on CLCXray. This trend is consistent with the proposed design where small prohibited items occupy limited spatial support and are easily obscured by surrounding structures, whereas the learned edge representation and early-stage boundary-guided corrections preserve local evidence that may be weakened in deeper contextual features. The BEF-Swin foundation already improves performance across all scales, while DERA provides further gains, particularly for small and medium objects, confirming that the boundary-residual pathway contributes beyond direct cross-stream fusion. Improvements for large objects are comparatively smaller, as their greater spatial extent already provides stronger semantic and geometric cues. Nevertheless, the positive AP$_L$ gains on all three datasets show that improved small-object sensitivity is achieved without sacrificing large-object detection.

\subsection{Ablation Studies} \label{subsec:ablations} 

\noindent All ablation studies are conducted exclusively on PIDray dataset. Each comparison uses fixed final checkpoints and matched training and evaluation settings. Results are reported using metrics detailed in Sec. \ref{subsec:metrics}.


\subsubsection{Scope of Swin Adaptation} \label{subsubsec:swin_scope}

\noindent Table~\ref{tab:swin_scope} examines whether X-ray adaptation can be confined to the later semantic stages of Swin while keeping the detection stack, PDC/PiDiNet branch, and edge-to-Swin projections trainable. Updating only S2--S3 preserves most of the full-hierarchy performance, with reductions of $0.3$ AP and $0.4$ AP$_{50}$/AP$_{75}$, suggesting that the early Swin representations remain largely transferable when the edge branch and fusion modules are adapted. However, this configuration removes only $1.5$M trainable parameters and therefore provides limited efficiency savings. Restricting adaptation to S3 yields a larger reduction in trainable parameters but decreases AP by $0.9$ and AP$_{75}$ by $1.1$. The stronger degradation at the stricter IoU threshold indicates that S2 contributes intermediate-resolution geometric information needed for precise box refinement, whereas S3 primarily captures high-level semantics.
\vspace{-0.4cm}
\begin{table}[ht]
\centering \caption{Effect of restricting foundation adaptation of Swin hierarchy stages. $\Delta$ is relative to our Bef-swin foundation adaptation results.} \label{tab:swin_scope} 
\scriptsize \setlength{\tabcolsep}{2.8pt} \renewcommand{\arraystretch}{1.12} \begin{adjustbox}{max width=\columnwidth} \begin{tabular}{@{}lcccc@{}} \toprule Adaptation scope & Trainable  & AP & AP$_{50}$ & AP$_{75}$ \\ \midrule \rowcolor{AblationBlue} \textbf{Bef-Swin} & \textbf{64.3M} & \textbf{74.5} & \textbf{85.7} & \textbf{79.9} \\ 
Later stages S2--S3 & 62.8M & 74.2\footnotesize{\textcolor{gray}{\textbf{\abdelta{$\Delta0.3\downarrow$}}}} & 85.3\footnotesize{\textcolor{gray}{\textbf{\abdelta{$\Delta0.4\downarrow$}}}} & 79.5\footnotesize{\textcolor{gray}{\textbf{\abdelta{$\Delta0.4\downarrow$}}}} \\ 
Final stage S3 only & 51.0M & 73.6\footnotesize{\textcolor{gray}{\textbf{\abdelta{$\Delta0.9\downarrow$}}}} & 84.8\footnotesize{\textcolor{gray}{\textbf{\abdelta{$\Delta0.9\downarrow$}}}} & 78.8\footnotesize{\textcolor{gray}{\textbf{\abdelta{$\Delta1.1\downarrow$}}}} \\ \bottomrule \end{tabular} \end{adjustbox}
\vspace{-0.5em}
\end{table}


\subsubsection{Edge branch study} \label{subsubsec:sobel_pdc}

\noindent Table~\ref{tab:sobel_pdc} isolates the effect of fixed and learnable edge extraction while preserving the same four-level hierarchy and fusion topology. The Sobel branch performs almost identically to the Grounding DINO baseline in AP and AP$_{50}$ (72.2 vs. 72.1, and 82.8 for both), but reduces AP$_{75}$ from 77.6 to 75.3. This indicates that generic image gradients provide little additional discriminative information and can degrade precise localization. In X-ray imagery, Sobel filters respond indiscriminately to material interfaces, overlapping benign objects, and baggage structures, allowing coarse detections to remain intact while perturbing the spatial support required for tighter boxes. This ablation shows that learned edge branch can adaptively emphasize task-relevant structure while suppressing irrelevant transitions.

\begin{table}[ht] 
\centering 
\vspace{-1em}
\caption{Comparison between Sobel edge branch and the learned pixel-difference branch. $\Delta$ is measured relative to our Bef-swin foundation adaptation results.} \label{tab:sobel_pdc} \scriptsize \setlength{\tabcolsep}{3.3pt} \renewcommand{\arraystretch}{1.12} \begin{adjustbox}{max width=\columnwidth} \begin{tabular}{@{}lcccc@{}} \toprule Edge branch & Trainable & AP & AP$_{50}$ & AP$_{75}$ \\ \midrule Sobel & $\sim 64.3M$ & 72.2\footnotesize{\textcolor{gray}{\textbf{\abdelta{$\Delta2.3\downarrow$}}}} & 82.8\footnotesize{\textcolor{gray}{\textbf{\abdelta{$\Delta2.9\downarrow$}}}} & 75.3\footnotesize{\textcolor{gray}{\textbf{\abdelta{$\Delta4.6\downarrow$}}}} \\ \rowcolor{AblationBlue} \textbf{PiDiNet} & $\sim $\textbf{64.3M}  & \textbf{74.5} & \textbf{85.7} & \textbf{79.9} \\ \bottomrule \end{tabular} \end{adjustbox} 
\vspace{-0.5em}
\end{table}

\subsubsection{Depth of the Boundary-Residual Pathway} \label{subsubsec:hierarchy_depth} 
\noindent Table~\ref{tab:hierarchy_depth} evaluates whether boundary prediction and residual injection remain beneficial when extended to lower-resolution hierarchy levels. Each variant trains fresh boundary heads, a matching fusion layer, and the corresponding zero-initialized residual projections from the same BEF-Swin foundation, thereby isolating the effect of pathway depth. The two-level design achieves the strongest performance despite using substantially fewer trainable parameters. Adding $P_2/S_2$ provides no gain and slightly reduces AP$_{75}$, while extending the pathway through $P_3/S_3$ causes a further decline as the residual capacity increases from 14.7K to 154.1K parameters. The progressively larger reduction at AP$_{75}$ suggests that deeper injection primarily affects localization precision as early stages retain the fine spatial detail needed for contour-guided refinement, whereas deeper features are coarser and increasingly semantic, making them less compatible with localized boundary cues and more susceptible to redundant or clutter-sensitive corrections.

\begin{table}[ht] 
\centering 
\vspace{-1em}
\caption{Effect of extending boundary prediction and residual adaptation through additional hierarchy levels. Parameter counts are shown separately for boundary-head learning and residual tuning. $\Delta$ is reported relative to DERA.} \label{tab:hierarchy_depth} \scriptsize \setlength{\tabcolsep}{2.3pt} \renewcommand{\arraystretch}{1.12} \begin{adjustbox}{max width=\columnwidth} \begin{tabular}{@{}lcccc@{}} \toprule Active hierarchy & Boundary / residual  & AP & AP$_{50}$ & AP$_{75}$ \\ \midrule \rowcolor{AblationBlue} \textbf{DERA: $P_0$--$P_1$ / $S_0$--$S_1$} & \textbf{26.1K / 14.7K} & \textbf{75.2} & \textbf{86.2} & \textbf{80.6} \\ 
Three levels: $P_0$--$P_2$ / $S_0$--$S_2$ &
60.8K / 61.2K &  75.1\footnotesize{\textcolor{gray}{\textbf{\abdelta{$\Delta0.1\downarrow$}}}} & 86.1\footnotesize{\textcolor{gray}{\textbf{\abdelta{$\Delta0.1\downarrow$}}}} & 80.3\footnotesize{\textcolor{gray}{\textbf{\abdelta{$\Delta0.3\downarrow$}}}} \\ 
Four levels: $P_0$--$P_3$ / $S_0$--$S_3$ &
95.4K / 154.1K & 75.0\footnotesize{\textcolor{gray}{\textbf{\abdelta{$\Delta0.2\downarrow$}}}} & 85.9\footnotesize{\textcolor{gray}{\textbf{\abdelta{$\Delta0.3\downarrow$}}}} & 80.2\footnotesize{\textcolor{gray}{\textbf{\abdelta{$\Delta0.4\downarrow$}}}} \\ \bottomrule \end{tabular} \end{adjustbox}
\end{table}

\subsubsection{Residual-Projection Initialization} \label{subsubsec:initialization} 

\noindent Table~\ref{tab:initialization} isolates the effect of residual initialization while keeping the boundary checkpoint, adapter placement, trainable parameter count, and optimization schedule unchanged. Standard and zero initialization achieve similar AP and AP$_{50}$, indicating that both configurations can learn useful residual corrections. However, zero initialization provides a clearer advantage at AP$_{75}$, improving it by 0.5 points. This suggests that preserving the foundation detector exactly at the start of adaptation is particularly beneficial for precise localization as the residual pathway is introduced without perturbing the established visual representation and gradually learns only detection-supported corrections. In contrast, standard initialization injects a non-zero response before training, which can slightly disturb early-stage spatial features.
\vspace{-0.3cm}

\begin{table}[ht] 
\centering \caption{Effect of residual-projection initialization. Both configurations use the same boundary checkpoint of DERA. $\Delta$ is reported relative to DERA.} \label{tab:initialization} \scriptsize \setlength{\tabcolsep}{3.5pt} \renewcommand{\arraystretch}{1.12} \begin{adjustbox}{max width=\columnwidth} \begin{tabular}{@{}lccccc@{}} \toprule Initialization & Initial correction & Trainable & AP & AP$_{50}$ & AP$_{75}$ \\ \midrule Standard initialization & Non-zero & 14.7K & 75.1\footnotesize{\textcolor{gray}{\textbf{\abdelta{$\Delta0.1\downarrow$}}}}  & 86.1\footnotesize{\textcolor{gray}{\textbf{\abdelta{$\Delta0.1\downarrow$}}}} & 80.1\footnotesize{\textcolor{gray}{\textbf{\abdelta{$\Delta0.5\downarrow$}}}}
\\ \rowcolor{AblationBlue} \textbf{Zero initialization} & \textbf{Exactly zero} & \textbf{14.7K} & \textbf{75.2} & \textbf{86.2} & \textbf{80.6} \\ \bottomrule \end{tabular} \end{adjustbox} 
\vspace{-2.5em}
\end{table}

\subsection{Qualitative Analysis}
\label{subsec:qualitative_analysis}

\begin{figure*}[!t]
\centering
\includegraphics[width=0.75\linewidth,height=8.5cm]{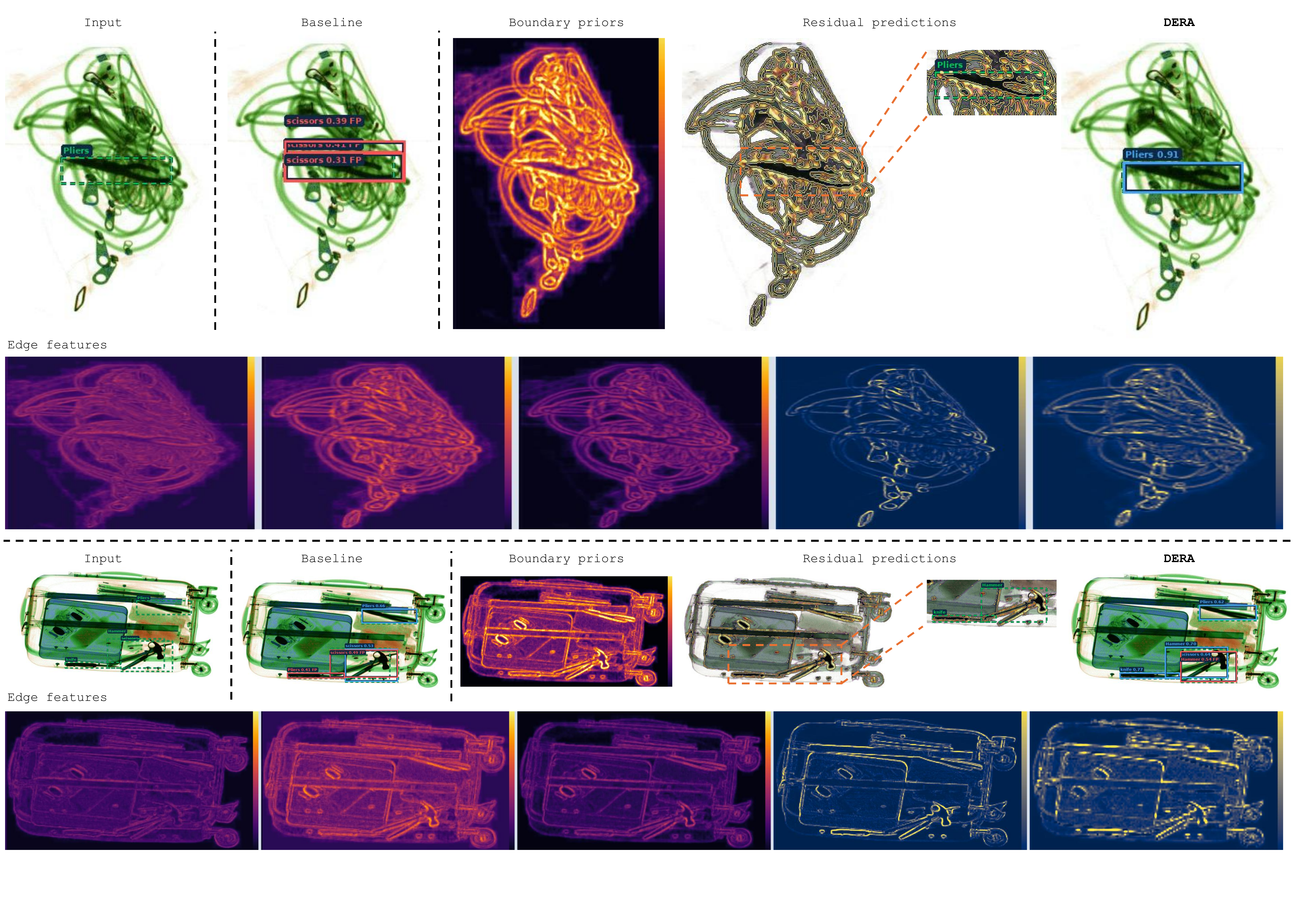}
\vspace{-2em}
\caption{Qualitative visualization of the baseline and DERA predictions, together with the learned boundary priors, residual responses, and multiscale edge features. Green annotations indicate ground truth, while red and blue boxes denote incorrect and correct detections, respectively.}

\label{fig:qualitative}
\vspace{-1em}
\end{figure*}

\noindent Figure~\ref{fig:qualitative} illustrates how DERA improves detection
under severe overlap and weak object appearance. In the shown examples, the
baseline produces incomplete or incorrect detections, whereas DERA recovers
the concealed targets with better-aligned bounding boxes. The
predicted boundary prior emphasizes object contours within the cluttered
baggage region, and the residual responses concentrate the resulting
correction around spatially relevant structures rather than uniformly
modifying the visual features. The edge-feature maps further show the complementary behavior of the
pixel-difference hierarchy where early levels preserve fine local transitions,
while deeper levels capture progressively coarser structural patterns.

\begin{figure*}[ht]
\centering
\includegraphics[width=0.75\linewidth,height=8.5cm]{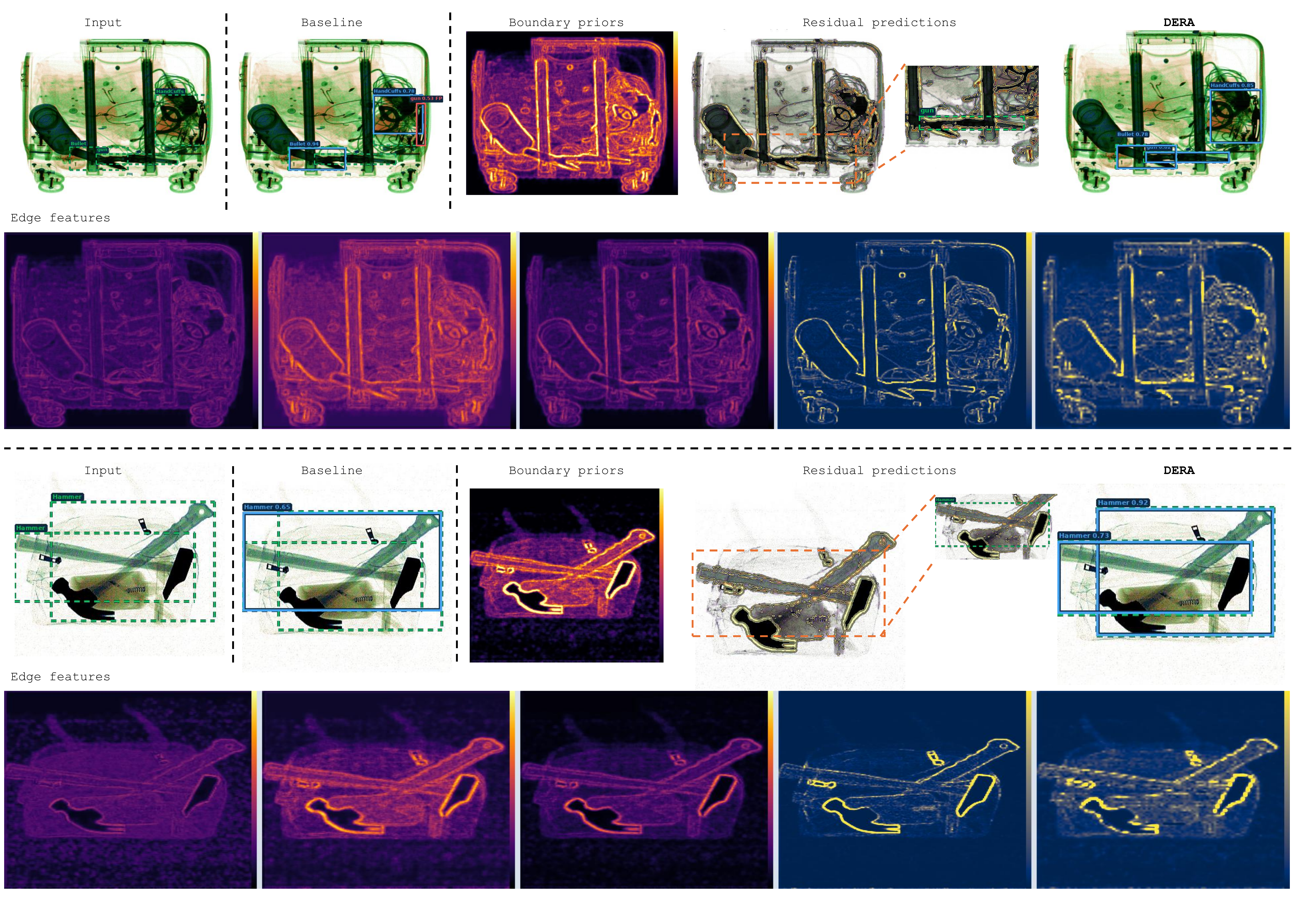}
\vspace{-0.7cm}
\end{figure*}

\begin{figure*}[ht]
\centering
\vspace{-1em}
\includegraphics[width=0.75\linewidth,height=8.5cm]{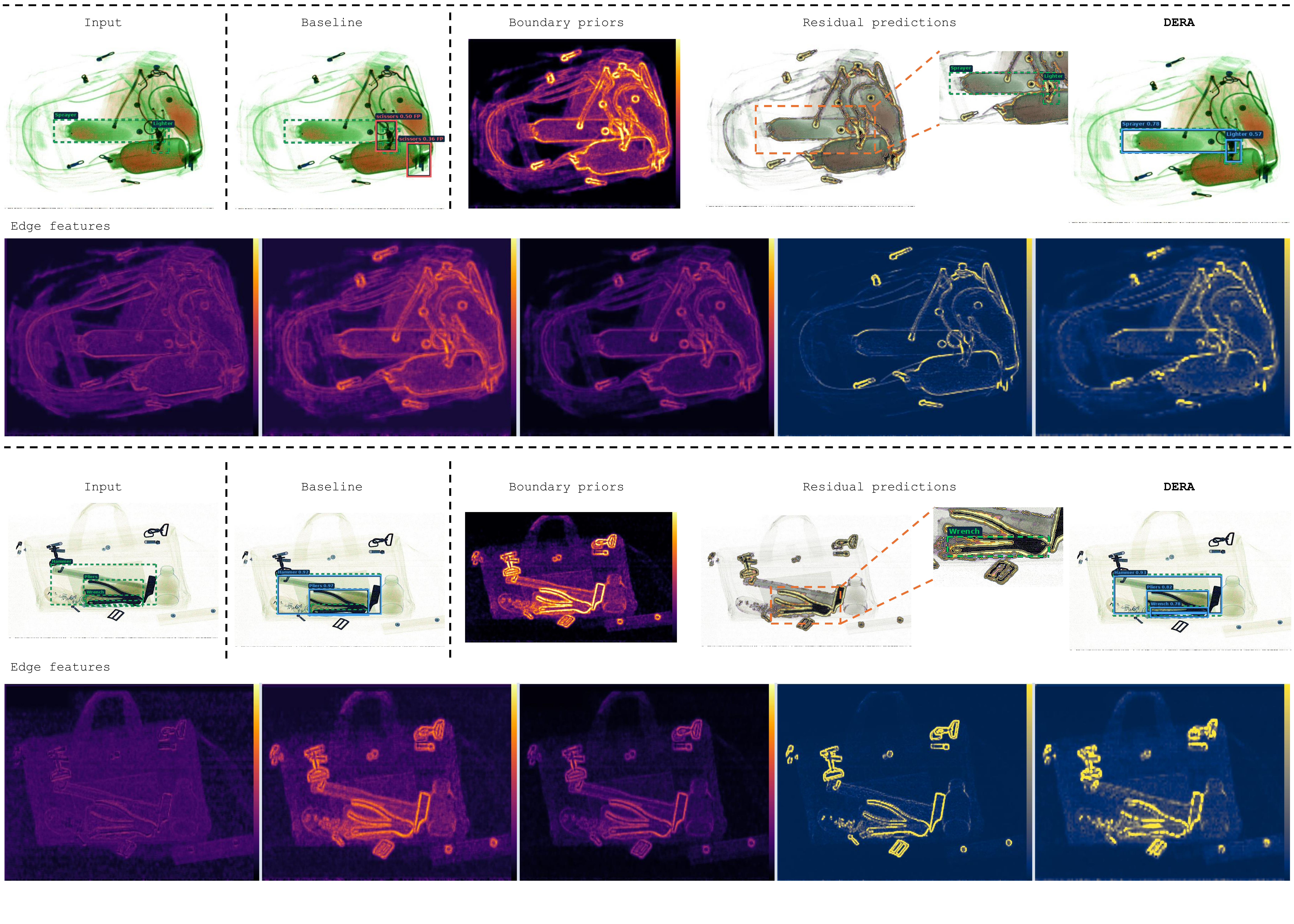}
\vspace{-0.7cm}
\end{figure*}
\begin{figure*}[!h]
\centering
\vspace{-1em}
\includegraphics[width=0.75\linewidth,height=8.5cm]{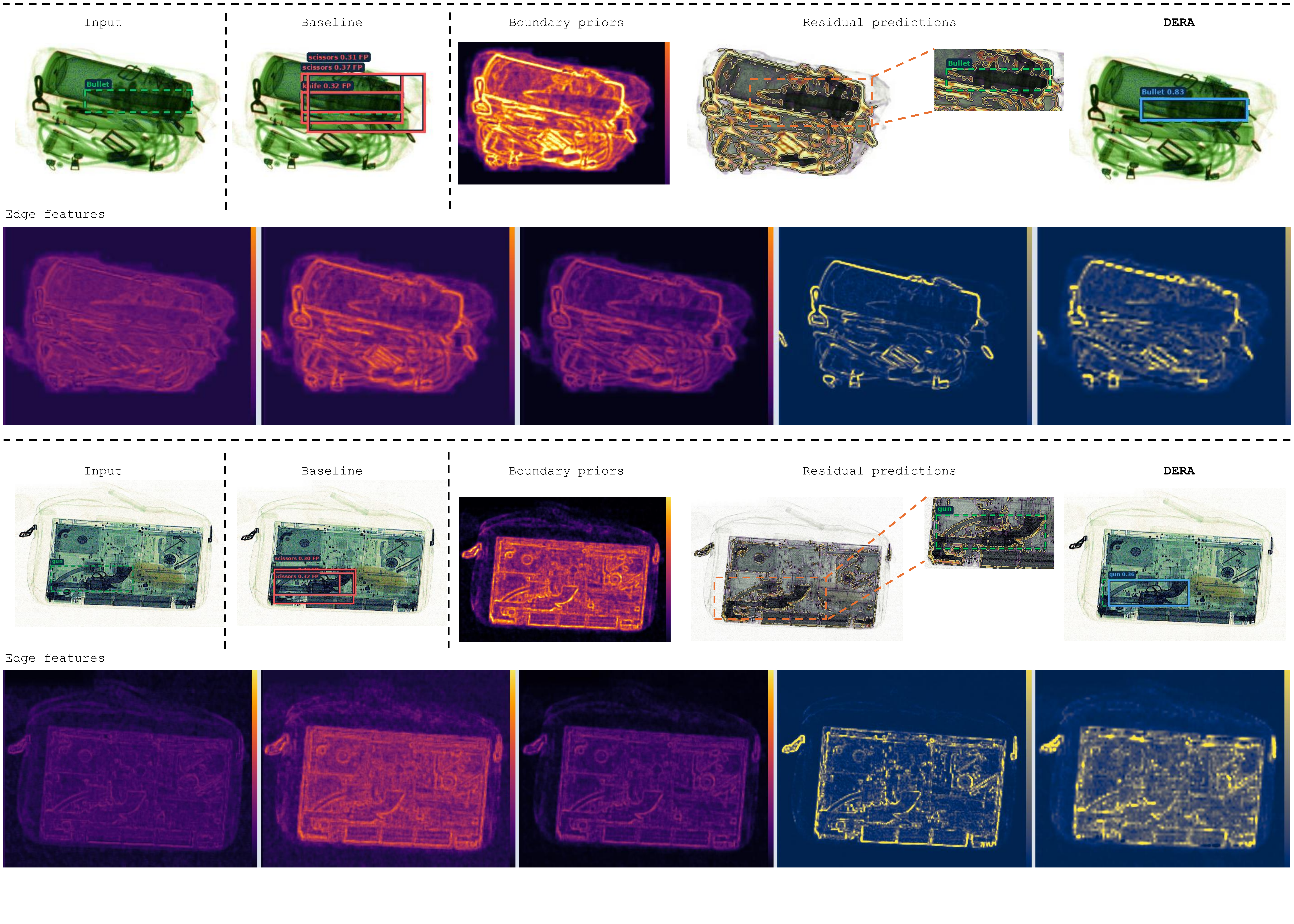}
\vspace{-1cm}
\end{figure*}

\subsection{Limitations and Failure Cases}
\label{subsec:failure_cases}

\begin{figure}[ht]
\vspace{-0.5em}
\centering
\includegraphics[width=0.99\linewidth]{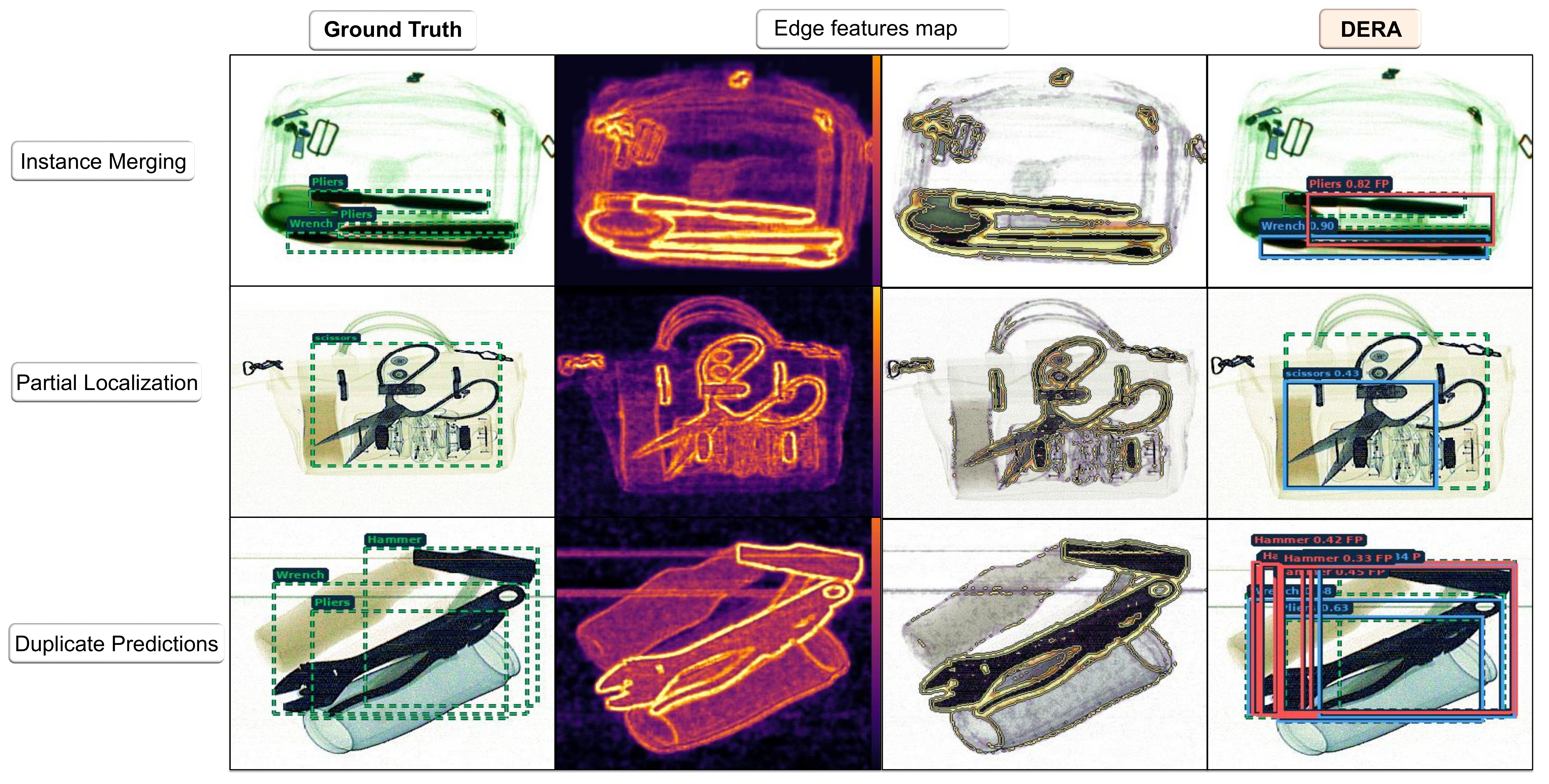}
\vspace{-2em}
\caption{\textbf{Failure cases of DERA.} Representative examples of instance
merging, partial localization, and duplicate predictions. Green dashed boxes
denote ground truth, while blue and red boxes indicate correct and incorrect
detections, respectively.}
\label{fig:failure}
\vspace{-1em}
\end{figure}

\noindent Figure~\ref{fig:failure} shows that boundary guidance alone does not
explicitly enforce instance-level separation or query consistency. Closely
packed objects may be merged into a single box, partially recovered objects
may remain insufficiently localized at strict IoU thresholds, and multiple
queries may respond to the same structure, producing duplicate detections.
These cases indicate that DERA improves structural representation but does not
fully resolve the downstream challenges of instance assignment, box
regression, and duplicate control.

\section{Conclusion} \label{sec:conclusion} 

\noindent This work presents DERA, an edge-aware framework for prohibited-item detection in X-ray imagery. DERA combines contextual and edge representations, learns an object-specific boundary prior from training-time instance contours, and injects boundary-guided corrections into the early visual stages through detached, residual heads. Experiments on PIDray, CLCXray, and STCray demonstrate consistent improvements across different imaging conditions, while the ablation studies validate the use of learned edge features, early-stage residual adaptation, and function-preserving initialization. DERA provides an effective and parameter-efficient adaptation strategy to improving detection under overlap, weak texture, and clutter.


\bibliographystyle{IEEEtran}
\bibliography{refs}


 



\vspace{-33pt}
\begin{IEEEbiography}[{\includegraphics[width=1in,height=1.25in,clip,keepaspectratio]{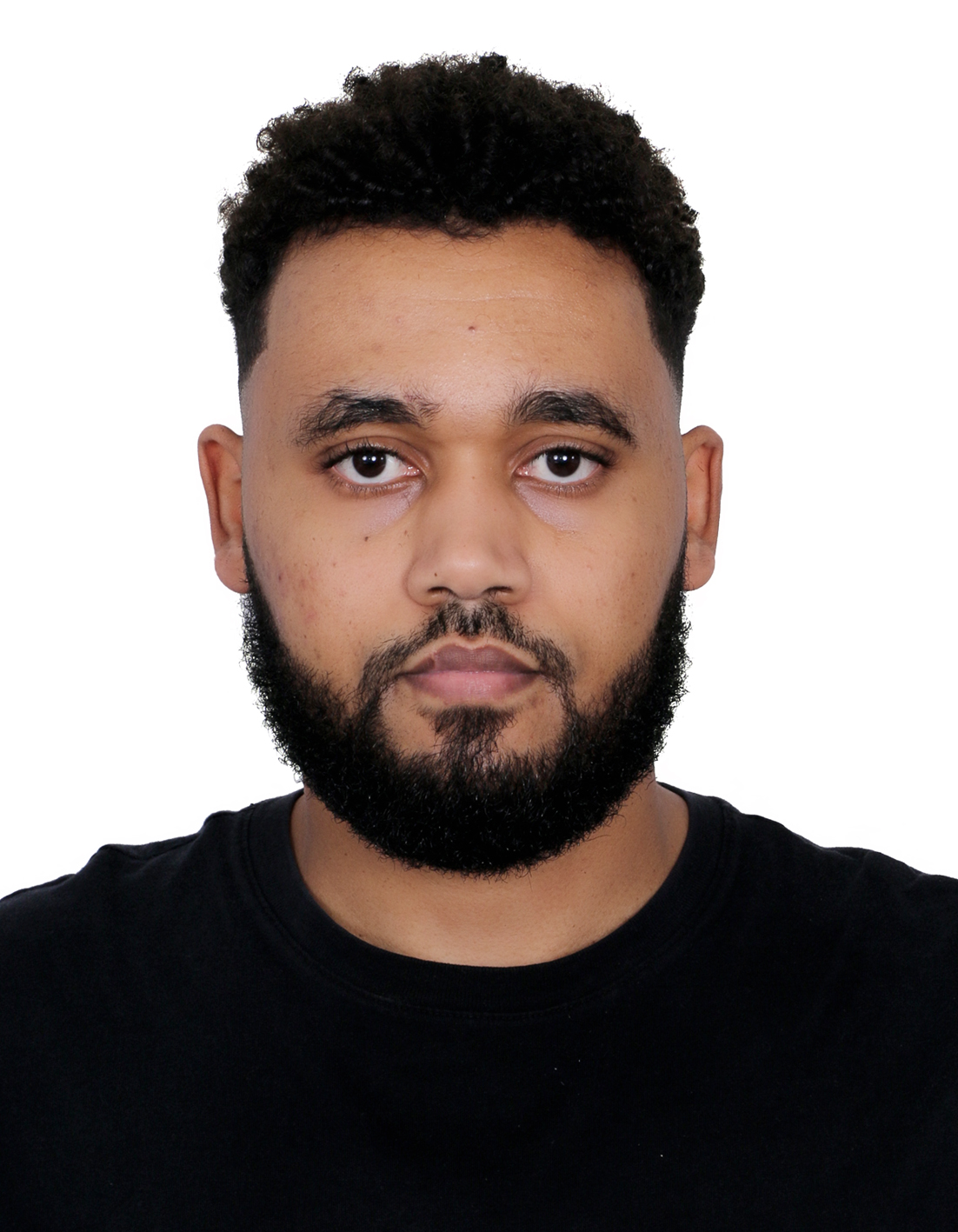}}]{Yonathan Michael}
received his B.Sc. in computer engineering from Khalifa University, where he is now pursuing his Ph.D. in the Department of Computer Science. His research interests include computer vision, multimodal large language models, and vision-language models, with a particular focus on X-ray security imaging and prohibited-item detection. 
\end{IEEEbiography}
\vspace{-1.5cm}
\begin{IEEEbiography}[{\includegraphics[width=1in,height=1.25in,clip,keepaspectratio]{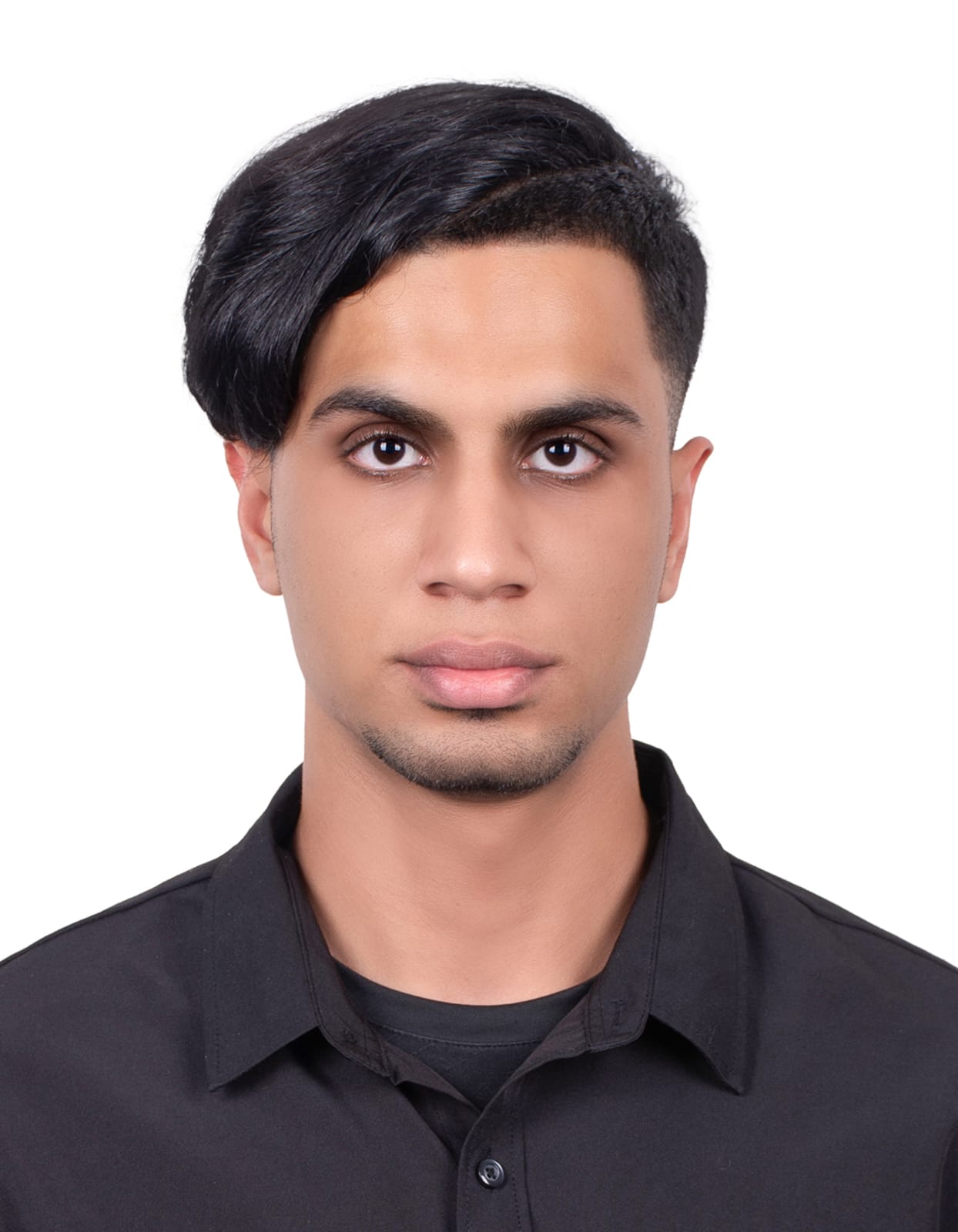}}]{Mohamad Alansari}
received his B.Sc. from Ajman University and his M.Sc. degree from Khalifa University in electrical and computer engineering, where he is currently pursuing his Ph.D. degree in computer science. His research interests include computer vision in autonomous robotics, multimodal large language models, and visual object tracking. 
\end{IEEEbiography}
\vspace{-1.5cm}
\begin{IEEEbiography}[{\includegraphics[width=1in,height=1.25in,clip,keepaspectratio]{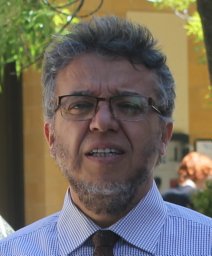}}]{Mohammed Bennamoun} (Senior Member, IEEE) is a Winthrop Professor of computer science at The University of Western Australia. He received his M.Sc. degree in control theory from Queen's University, Canada, and his Ph.D. degree in computer vision in Australia. His research interests include computer vision, 3D vision, deep and multimodal learning, biometrics, image processing, and robotics. He has authored four books and published more than 540 journal and conference papers.
\end{IEEEbiography}
\vspace{-1.5cm}
\begin{IEEEbiography}[{\includegraphics[width=1in,height=1.25in,clip,keepaspectratio]{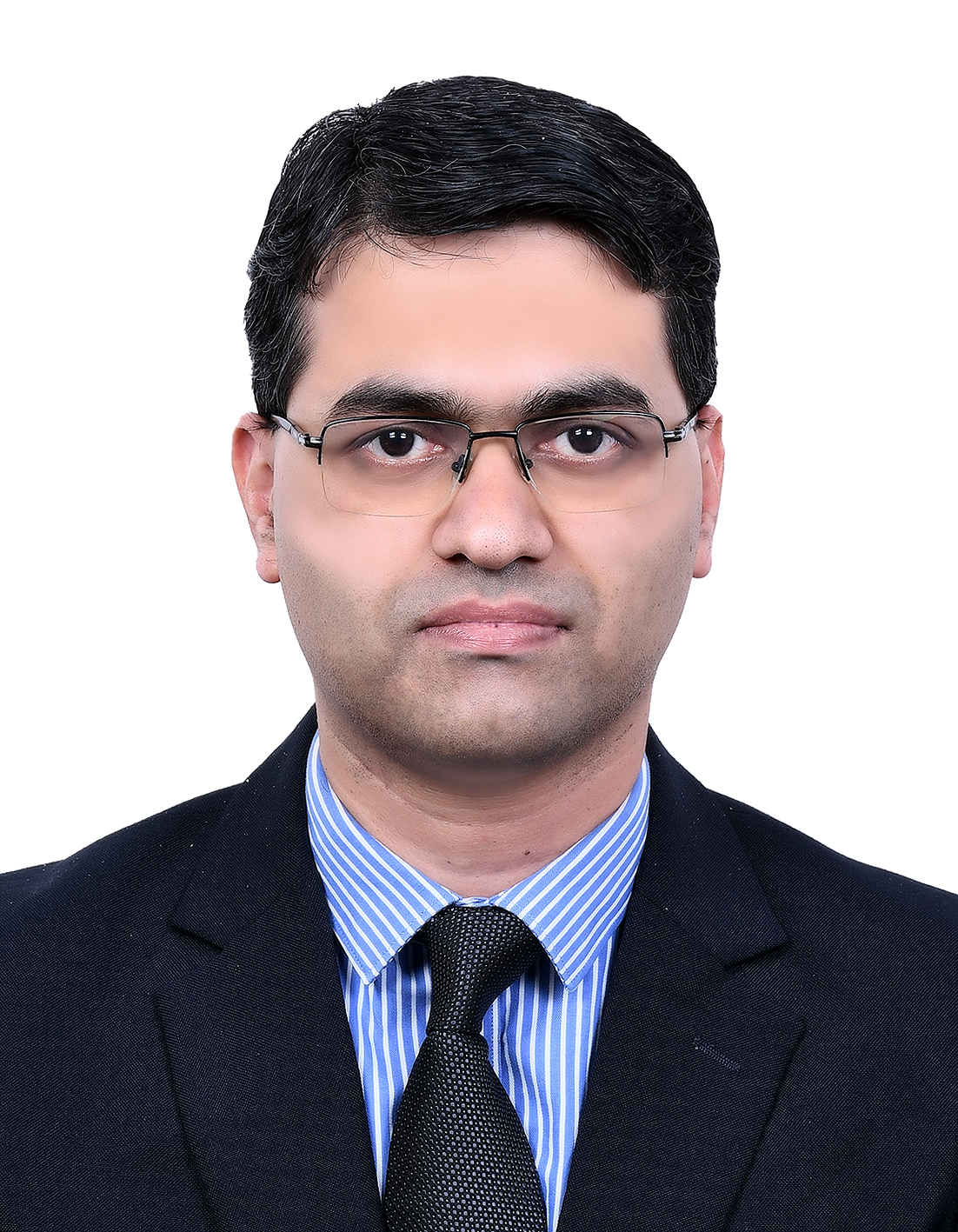}}]{Dwarikanath Mahapatra}
is an Assistant Professor in the Department of Computer Science at Khalifa University. He received his Ph.D. from the National University of Singapore and subsequently worked as a postdoctoral research fellow at ETH Zurich and a Research Staff Member at IBM Research Australia. His research interests include medical image analysis, machine and deep learning, computer-aided diagnosis, object detection, tracking, and image classification. He has published more than 100 research papers and holds 15 patents.
\end{IEEEbiography}
\vspace{-1.5cm}
\begin{IEEEbiography}[{\includegraphics[width=1in,height=1.25in,clip,keepaspectratio]{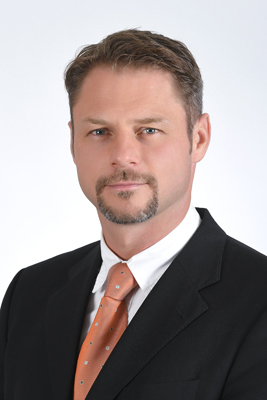}}]{Andreas Henschel}
received his M.Sc. and Ph.D. degrees in computer science from the Technical University of Dresden, Germany, in 2002 and 2008, respectively. He subsequently joined Masdar Institute as a postdoctoral researcher and later became an Assistant Professor. He was also a Visiting Scholar at the Massachusetts Institute of Technology. His research interests include bioinformatics, artificial intelligence, data science, genomic analysis, deep learning, and transfer learning, with applications to microbiome analysis, genomic epidemiology, and genotype-phenotype prediction.
\end{IEEEbiography}

\begin{IEEEbiography}[{\includegraphics[width=1in,height=1.25in,clip,keepaspectratio]{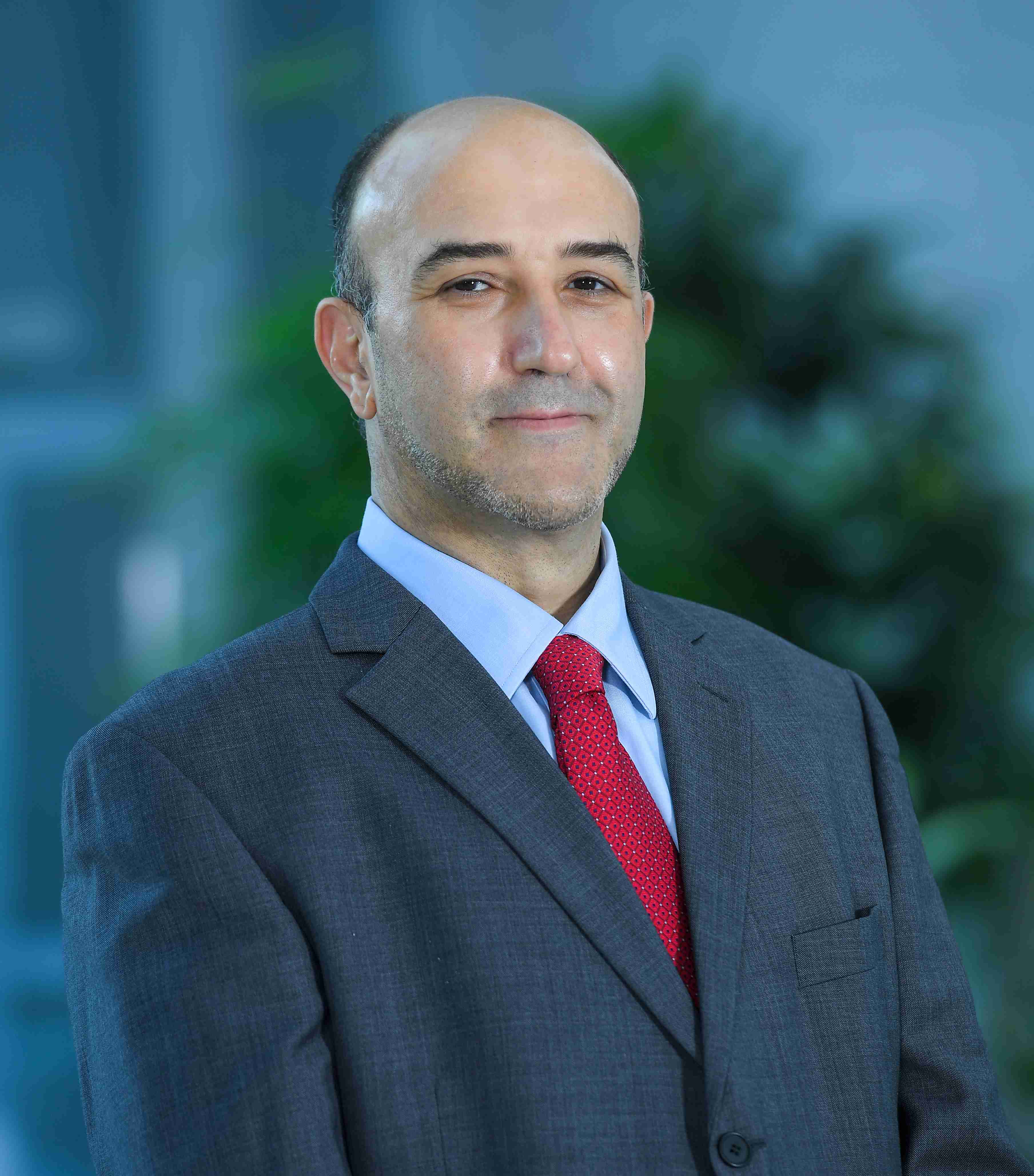}}]{Naoufel Werghi} (Senior Member, IEEE) is a Full Professor in the Department of Electrical Engineering and Computer Science at Khalifa University. He received his Habilitation and Ph.D. degrees in computer vision from the University of Strasbourg and previously held academic positions at the Universities of Edinburgh and Glasgow. His research interests include computer vision, machine learning, biometrics, medical imaging, remote sensing, surveillance, and intelligent systems. He has secured 23 research grants, published more than 200 journal and conference papers, and received five best paper awards. He serves as an Associate Editor of the IEEE Transactions on Circuits and Systems for Video Technology and the EURASIP Journal on Image and Video Processing.
\end{IEEEbiography}

\end{document}